\documentclass[pdflatex,sn-basic, iicol]{sn-jnl}

\usepackage{graphicx}%
\usepackage{multirow}%
\usepackage{amsmath,amssymb,amsfonts}%
\usepackage{amsthm}%
\usepackage{mathrsfs}%
\usepackage[title]{appendix}%
\usepackage{xcolor, colortbl}%
\usepackage{textcomp}%
\usepackage{manyfoot}%
\usepackage{booktabs}%
\usepackage{algorithm}%
\usepackage{algorithmicx}%
\usepackage{algpseudocode}%
\usepackage{listings}%
\usepackage[capitalize]{cleveref}
\usepackage{makecell} 
\usepackage{multirow}
\usepackage{hhline}
\usepackage{pifont}
\usepackage{diagbox}
\usepackage{enumitem}
\usepackage{tablefootnote}
\usepackage{float}
\usepackage{adjustbox}
\usepackage{ragged2e} 
\usepackage[T1]{fontenc}
\usepackage{mathptmx}

\newcommand{\cmark}{\ding{51}}
\newcommand{\xmark}{\ding{55}}
\definecolor{lightergray}{cmyk}{0,0.0,0.0,0.14}
\newcolumntype{?}{!{\vrule width 2pt}}

\newcommand{\lowformeratt}{Lowtention}

\theoremstyle{thmstyleone}%
\theoremstyle{thmstyletwo}%

\theoremstyle{thmstylethree}%

\begin{document}

\title[Article Title]{Beyond MACs: Hardware Efficient Architecture Design\\ for Vision Backbones}


\author*[1]{\fnm{Moritz} \sur{Nottebaum}}\email{nottebaum.moritz@spes.uniud.it}

\author[1,2]{\fnm{Matteo} \sur{Dunnhofer}}\email{matteo.dunnhofer@uniud.it}

\author[1]{\fnm{Christian} \sur{Micheloni}}\email{christian.micheloni@uniud.it}

\affil[1]{\orgdiv{ Machine Learning and Perception Lab}, \orgname{University of Udine}

\orgaddress{\street{Via delle Scienze 206}, \city{Udine}, \postcode{33100}, \state{UD}, \country{Italy}}}

\affil[2]{\orgdiv{ Centre for Vision Research}, \orgname{York University}

\orgaddress{\street{4700 Keele St}, \city{Toronto}, \postcode{M3J 1P3}, \state{ON}, \country{Canada}}}

\abstract{
Vision backbone networks play a central role in modern computer vision. Enhancing their efficiency directly benefits a wide range of downstream applications. 
To measure efficiency, many publications rely on MACs (Multiply Accumulate operations) as a predictor of execution time.
In this paper, we experimentally demonstrate the shortcomings of such a metric, especially in the context of edge devices. By contrasting the MAC count and execution time of common architectural design elements, we identify key factors for efficient execution and provide insights to optimize backbone design.
Based on these insights, we present LowFormer, a novel vision backbone family.
 LowFormer features a streamlined macro and micro design that includes \lowformeratt, a lightweight alternative to Multi-Head Self-Attention. \lowformeratt\ not only proves more efficient, but also enables superior results on ImageNet.
 Additionally, we present an edge GPU version of LowFormer, that can further improve upon its baseline's speed on edge GPU and desktop GPU.
  We demonstrate the LowFormer's wide applicability by evaluating it on smaller image classification datasets, as well as adapting it to several downstream tasks, such as object detection, semantic segmentation, image retrieval, and visual object tracking.
 LowFormer models consistently achieve remarkable speed-ups across various hardware platforms compared to recent state-of-the-art backbones.
 Code and models will be made publicly available. 
}


\maketitle

\section{Introduction}
\label{sec:intro}

\begin{figure}[hbt!] 
  \includegraphics[width=\linewidth]{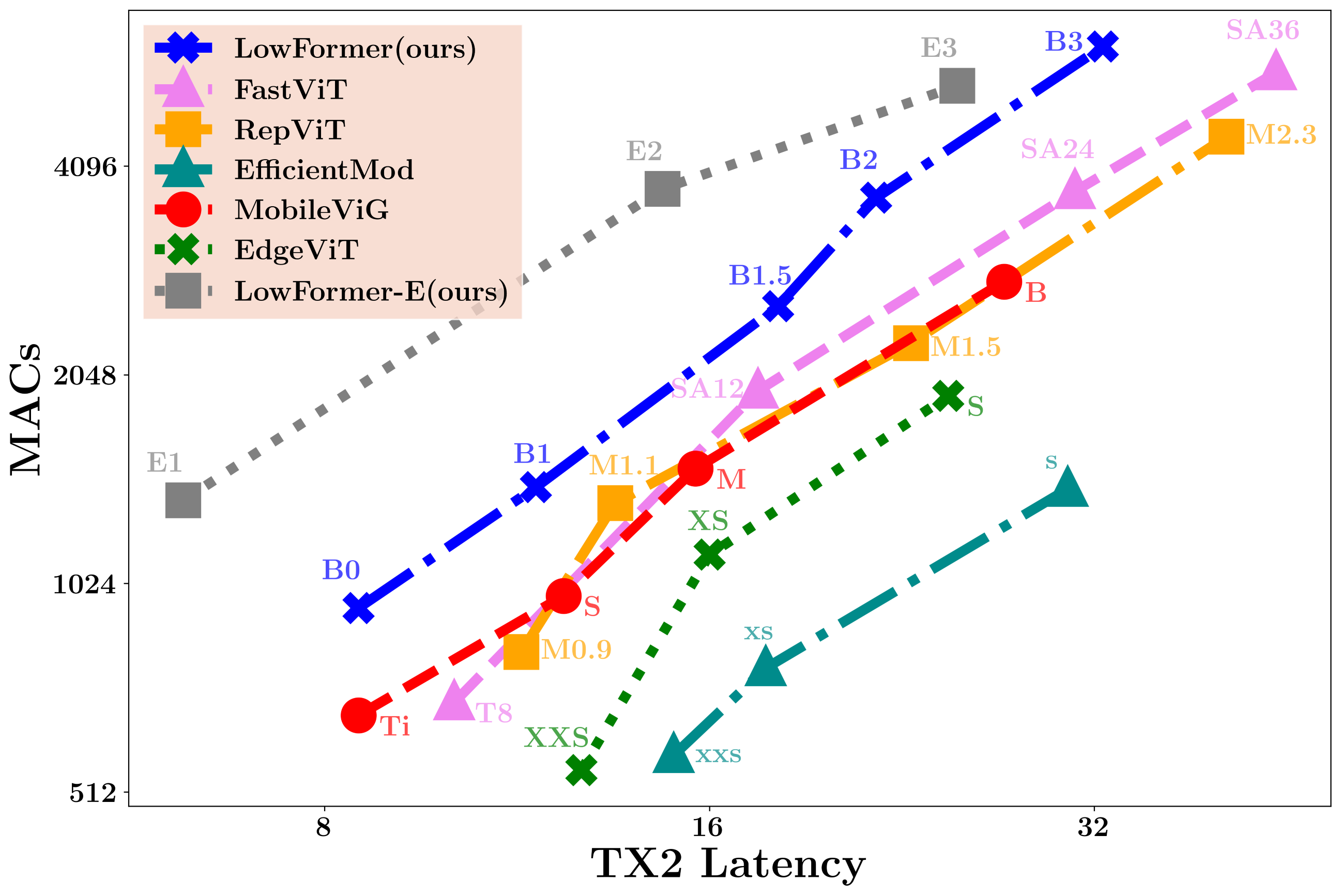}
  
  \caption{Comparison of hardware efficiency of different vision backbone architecture families  on the Nvidia Jetson TX2. Models in the top-left offer the best hardware efficiency on the Jetson TX2. Both axes are in logarithmic scale. 
LowFormer base models (B0-B3) outperform all architectures in hardware efficiency, with edge variants (E1-E3) further enhancing efficiency}
  \label{fig:hardwareefficiency}

\end{figure}

In many computer vision applications, it is critical to achieve accurate predictions in the shortest time possible. This is  important in real-time domains such as robotics \citep{efficientrobotics,dunnhofer2021weakly,matthies2007computer}, autonomous driving \citep{efficientautonomousdriving,jiang2023vad}, video surveillance \citep{efficientvideosurveillance,upmanyu2009efficient}, and user assistance  \citep{efficientassistancesystems,tan2023egodistill,leo2017computer}, and especially when these systems have to be deployed on mobile and edge devices \citep{mobileone,yolo-retedge}.

Nowadays, vision backbone networks are critical components of such  computer vision systems. They are used to generate representations that support a wide
range of high-level tasks for image \citep{Detr,segmentanything,groundingdino,ma2021image} and video
understanding \citep{Stark,matteovisobjtracking,kong2022human}. Improving the efficiency of vision
backbones is, therefore, a crucial step towards enhancing the running time of many computer vision pipelines.

Since the introduction of deep convolutional  networks \citep{lecun1989backpropagation}, vision backbones have evolved to balance accuracy and efficiency. Recent architectures \citep{fastvit, efficientmodulation, biformer, mobileformer} combine convolutional \citep{mobilenetv2,resnetpaper} and attention layers \citep{hydraattention, attentionisallyouneed, linformer}: convolutions extract local image features, while attention captures global relationships by aggregating information across the image. 
To develop computationally efficient deep learning models, including vision backbones,  researchers commonly count and minimize Multiply and Accumulate operations (MACs) \citep{biformer}. Simply put, this metric counts the
number of multiplications and additions performed by a neural network  to compute the output from
input data. In other words, MACs can be viewed as a measure of the ``tasks'' a model must perform to produce an output. Generally, the fewer tasks required, the faster and more efficient the model will be \citep{efficientnet}. But research has also shown a strong correlation between a model’s increased MAC count and its accuracy, with networks having higher MAC counts achieving better prediction accuracy \citep{efficientnet,scalingvits}. 
In light of this evidence, the research community has increasingly focused on developing backbone architectures that maximize accuracy and minimize the number of MACs.
By doing so, it is claimed that the accuracy-speed trade-off is optimized.

We argue that MAC count is not always the best measure of backbone efficiency, and the reduction of MACs does not necessarily translate to a backbone that achieves reduced execution time. 
This disproportion stems from factors like memory access costs and the degree of
parallelism \citep{efficiencymisnomer,efficientvitmemory, mobileone, hardwareffficientdesign}. 
The former identifies the delay caused by an operation that must wait for the retrieval of its operands from the memory, leading to idle periods that hinder overall execution time \citep{fasternet}.
The latter refers to the parallel execution of core operations like multiplications and additions on modern hardware. In these settings, the number of MACs still counts all the tasks required to produce the output, but fails to account for time saved by performing multiple operations simultaneously.

As a result, models vary in their hardware efficiency, leading to differences in execution time for the same amount of MAC operations.
In \Cref{fig:hardwareefficiency}, we analyze the hardware efficiency of different vision backbone families, by relating MAC count and actual execution time -- referred to as latency -- on an Nvidia Jetson TX2.
The figure depicts significant differences in hardware efficiency between the compared vision backbone architecture families. The MAC count of models within an architecture family (e.g. LowFormer) correlates well with execution time (TX2 latency), meaning they feature a similar hardware efficiency. However, the hardware efficiency differs significantly between the architecture families. The EfficientMod \citep{efficientmodulation} models (xxs, xs and s) for example have a similar latency as LowFormer-B1.5, B2 and B3, but a fraction of their MAC operations.
This emphasizes the importance of a sound architecture design, as a
measure to improve hardware efficiency.

Determining the extent to which a model's architecture influences hardware efficiency,
 due to memory access cost and level of parallelism, is complex and heavily dependent on hardware-specific implementation details \citep{mobilenetv4}.
We believe that latency is a more effective metric for evaluating vision backbone efficiency, as model efficiency cannot be determined by MAC count alone but must be tested across various execution devices.

Many recent publications have contributed to improve the understanding of the relationship between MACs and execution time on real devices \citep{mobileone, fasternet, shufflenetv2, repvit}, highlighting architectural configurations that deteriorate model speed.
In this paper, we add to this knowledge by pointing out additional factors, like: the operating resolution for convolutions;
the shortcomings of the frequently applied mobile inverted bottleneck block \citep{mobilenetv2}; 
and the considerable inefficiency of Multi-Head Self-Attention (MHSA) \citep{attentionisallyouneed}.
All of these factors can slow down model execution, however in what magnitude depends on the hardware device, wherefore we exploit a variety of different 
devices for a new hardware efficiency study.

\begin{figure}[bt!] 
  \includegraphics[width=\linewidth]{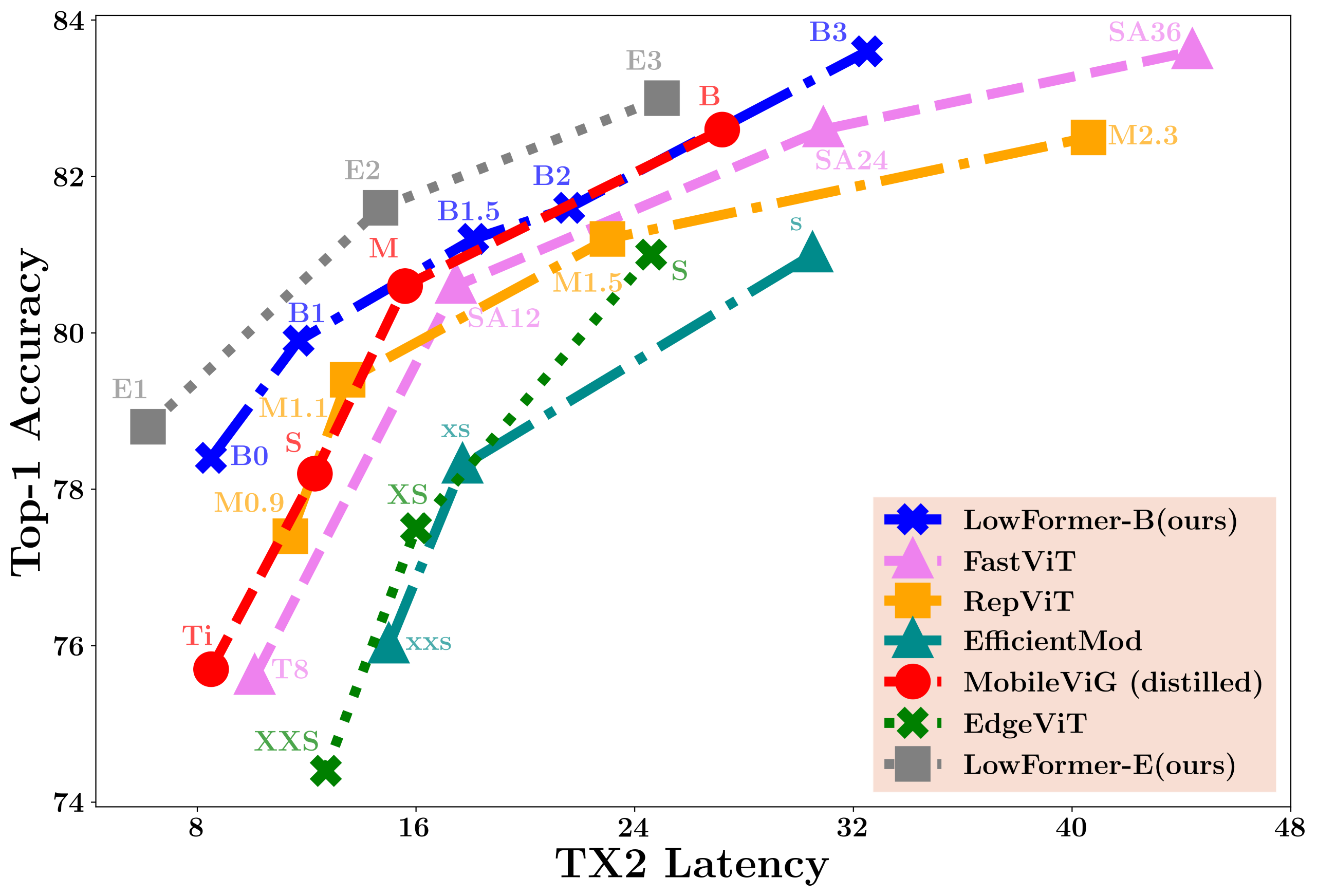}
  
  \caption{Comparison of Nvidia Jetson TX2 latency and top-1 accuracy for state-of-the-art vision backbones with LowFormer. Models in the top-left offer the best speed-accuracy trade-off. LowFormer consistently achieves lower latency for similar accuracy. Its edge variants (E1/E2/E3) further enhance this trade-off over the base models (B0-B3)  }
  \label{fig:throughputcompfig}

\end{figure}

We rely on the outcomes of such a study
to develop LowFormer, a new class of vision backbone networks that mitigates the impact of the aforementioned factors.
By minimizing  execution time instead of MAC count, LowFormer is able to achieve a new position in the state-of-the-art speed-accuracy trade-off (see \Cref{fig:throughputcompfig}), exceeding previous approaches on a variety of hardware devices. 
A key component in LowFormer's architecture is  \lowformeratt, a new and lightweight adaptation of MHSA \citep{attentionisallyouneed}.
LowFormer features a simple micro and macro design, enabling scalability
from low model model size (LowFormer-B0) to higher (LowFormer-B3).
The backbone family includes five models (B0, B1, B1.5, B2, B3), achieving top-1 accuracy on the ImageNet-1K \citep{imagenet} dataset ranging from 78.4\% to 83.64\%.
We further extend the architecture family by three models, namely LowFormer-E1/E2/E3, which are specifically targeted for edge  GPU devices and are derivations of the original LowFormer design. 
To confirm the applicability of LowFormer backbones in downstream tasks, we evaluated LowFormer's transfer learning capabilities on several image classification datasets. Furthermore,
we integrated our backbones in  object detection and semantic segmentation frameworks. Additionally, we utilize the  embeddings of LowFormer backbones for image retrieval, and present a LowFormer-based visual object tracking architecture. 
In all of these applications, LowFormer contributes significantly to improve hardware efficiency while maintaining or even increasing accuracy.

The contributions of this paper can be summarized as follows:
\begin{itemize}

    \item We carry out a new exhaustive hardware efficiency analysis of key design elements in vision backbone architectures. We show how the MAC count of those elements translates to execution time on several devices, and highlight their differences for overall model efficiency.         
     \item We present LowFormer, a new family of vision backbones that features a hardware-efficient macro design and a new lightweight
            attention operation. These backbone models are faster in terms of latency and throughput
            compared to models with similar accuracy.
    \item We further propose three edge GPU variants of LowFormer, that improve upon the base models in terms of edge GPU efficiency. 
    \item We show that LowFormer models generalize well to several downstream tasks, such as image classification, object detection, semantic segmentation, image retrieval, and single object tracking. 

\end{itemize}


\section{Related Work}
\label{sec:relatedwork}

\subsection{Hardware-Efficient Model Design}
\label{subsec:relatedwork_efficientmodeldesign}

Achieving the highest accuracy at all computational cost has long since ceased to be the only goal in vision backbone networks \citep{efficientnet}. An ever increasing share of research focuses on developing the most efficient architecture, subsequently achieving the best accuracy-speed trade-off \citep{edgevit, mobilevig, mobilenetv3}. Earlier approaches equated speed with the minimal amount of MACs \citep{ghostnetv2,ghostnetv1,mobileformer}, while more recent research increasingly aims for models that reduce latency or throughput -- i.e. number of images processed in a second -- on several different types of hardware such as desktop GPU and CPU \citep{shufflenetv2,shvit,efficientvit}, mobile NPU and GPU \citep{mobileone}, or microcontroller CPU \citep{phinetsembeddingdevice}.
Within these approaches, the cost of memory access and the degree of hardware parallelism have become important factors for efficient model design, as they can have a significant impact on model speed \citep{fasternet, mobilenetv4}.
The design of EfficientViT \citep{efficientvitmemory} demonstrates that MHSA introduces a higher memory access cost compared to the Feed-Forward Network (FFN) within the transformer block. Therefore, this architecture proposes increasing the proportion of FFN operations relative to MHSA, resulting in improved efficiency without compromising accuracy. 
MobileOne \citep{mobileone}, on the other hand, is based on an analysis of how activation functions and multi-branch architectures impact latency on mobile devices. 
ShuffleNetV2 \citep{shufflenetv2} and FasterNet \citep{fasternet}  were proposed on the observation that grouped convolutions are executed inefficiently on GPUs due to their high memory access costs \citep{fasternet}.

In this paper, we draw inspiration from these insights. However, rather than ungrouping a portion of the convolutions  \citep{shufflenetv2} or introducing a new micro design \citep{fasternet}, we study the impact of fusing depthwise and pointwise convolutions on execution time.
Additionally, we examine how the operating resolution of convolutional layers impacts latency, and further explore strategies to effectively mitigate the significant efficiency drop caused by increasing input resolution for MHSA.
To the best of our knowledge, we are the first to assemble a diverse set of execution devices to perform these efficiency experiments and analyze how different hardware platforms compare.

This paper extends \citep{nottebaum2024lowformer}, where the LowFormer architecture was initially presented. In this version, we provide additional contributions, tailored to the domain of edge computing.
We expand the execution time analysis to cover several edge devices and examine the efficiency of MHSA for increased input resolution, exploring ways to improve it.
We also consistently compare the efficiency of LowFormer on edge devices with the top competing models, across most benchmarks.
We further extend the ablation study to provide a stronger rationale for our design choices.
Additionally, we propose three edge GPU variants of LowFormer and demonstrate empirically their viability.
We evaluate LowFormer on several new downstream tasks, including image classification and image retrieval. 
Lastly, we present LowFormer-Track, which improves performance of the SMAT \citep{smattracker} single object tracking (SOT) framework by integrating LowFormer's design principles.

\subsection{Convolutions, Attention and MLP in Architecture Design}
Most modern backbone architectures consist entirely or partially of three main building blocks: convolutions, attention mechanisms, and multi-layer perceptrons (MLPs) \citep{shvit,fastvit}. Some approaches, called hybrid models, 
combine all three in their design \citep{efficientvitmemory,mobilenetv4,shvit, fastvit}, some relieve of the MLP \citep{efficientvit}, while others solely rely on convolutions \citep{repvit, mobileone}. The latter
focus towards efficient mobile execution, where convolutions have been shown to achieve superior latency results \citep{mobilenetv4}.
The work of \citep{nfnetconvonlybatchnorm} demonstrates that purely convolutional backbones can be on par with hybrid models in terms of accuracy, however \citep{CoatNet} experimentally show that the attention operation provides higher model capacity if incorporated in an architecture. 
On the other side, convolutions exhibit an improved generalization ability compared to attention modules,
wherefore a combination of both on a macro design level is beneficial \citep{CoatNet}.
Other works went further by joining convolutions and attention operations on a micro design level \citep{CvT}, relieving of the need for hand-crafted positional encoding as a result.
 
In the design of the LowFormer architecture, we integrate all three building blocks (attention, convolution, MLP) in a straightforward 
manner, resulting in a robust and versatile architecture. Unlike other approaches that depend on neural architecture search \citep{mobilenetv4,efficientnet} resulting in irregular macro designs, enforce a fixed micro design \citep{efficientvitmemory}, or entirely exclude MLPs and/or attention mechanisms to improve performance \citep{mobileone, repvit, efficientvit}, LowFormer takes a more flexible approach. Its architecture allows the removal of any building block without compromising its core design principles.
In \Cref{subsec:adapting_lowformer_for_edge}, we present a variant of the original LowFormer architecture, where we remove a portion of said components to further boost efficiency on edge GPU devices.

\subsection{Efficient Attention}
The landscape of attention mechanisms is vast, with many alternatives proposed to replace MHSA \citep{attentionisallyouneed}.
A lot of effort has been spent  \citep{hydraattention,efficientvit, maxvit} to reduce its quadratic complexity by variations of linear attention \citep{linformer}. 
On the other side, the work \citep{poolformer} shows that attention can be replaced by a simple pooling operation.
Subsequent research \citep{efficientformer} takes that idea further and uses the efficient pooling operation for the first three backbone stages and the traditional attention for the last two stages \citep{efficientformer}. 
Other works \citep{CvT,PyramidTransformer,multiscalevit} downsample the key and value vectors before the attention operation, either with convolutions or pooling. 
Pooling is also used to downsample all query, key, and value vectors in order to make attention completely operate on a lower resolution \citep{inceptiontransformer}.

In contrast to previous works, we harness the learning capability of convolutions to downscale the resolution of all input vectors for the Scaled Dot-Product Attention (SDA), effectively serving as conditional position embeddings \citep{convposencoding}. Unlike others, we further reduce the channel dimension before the SDA.
Both reductions -- of resolution and channel dimension -- have a significant effect on efficiency 
but a minimal effect on accuracy.

\begin{table*}[h!]
    \centering
    \caption{Contrasting MAC count and execution time of depthwise convolutions and ungrouped convolutions} 
    \label{tab:groupingexp}
    \begin{threeparttable}[b]
        
    \resizebox{1.0\textwidth}{!}{\begin{tabular}{c|c|c|c|c|c|c|c|c}
        \Xhline{1.5pt}
         \multirow{2}{*}{Model} & Channel & \multirow{2}{*}{Depthwise} & MACs &  GPU Throughput $\uparrow$ & \multicolumn{4}{c}{Latency(ms) $\downarrow$ }  \\ 
         
                  & {[C$_0$, C$_1$, C$_2$, C$_3$, C$_4$]} & &  (M) & (images/s) & \multicolumn{1}{c}{TX2 } & \multicolumn{1}{c}{GPU } & \multicolumn{1}{c}{Mobile } & \multicolumn{1}{c}{ARM CPU } \\
         
         \Xhline{1.0pt}
        
         \#1 & {[15, 30, 60, 120, 240]}  & \xmark & 463 &  \textbf{12,722}  & \textbf{1.83} & 0.78 &  0.63 &  \textbf{12.58} \\
         \#2 & {[30, 60, 120, 240, 480]} & \cmark \par & 42  & 10,526  & 3.55 & \textbf{0.24 }  & \textbf{0.53}  & 17.71  \\
         \Xhline{1.0pt}
        
         \#3 & {[30, 50, 100, 160, 160]}  & \xmark & 956  & \textbf{7,142} & \textbf{2.90} & 0.78   &  \textbf{0.64} & \textbf{24.29}  \\
        \#4 & {[60,120,240,480,480]}  & \cmark \par & 82  & 5,422  &  6.16 & \textbf{0.39}    & 0.67 &  36.59  \\
        
        \Xhline{1.0pt}
        
         \#5 & {[30, 60, 150, 240, 240]}  & \xmark & 1710  & \textbf{5,350}  & \textbf{4.27} & 0.99   & 1.00  & 48.19    \\
         \#6 & {[60, 180, 360, 720, 720]} & \cmark \par & 104 & 4,244  &  8.03 & \textbf{0.43}   & \textbf{0.86}  &  \textbf{47.29}   \\

        \Xhline{1.5pt}

    \end{tabular}}

\begin{tablenotes}[para,flushleft]
\footnotesize 
    The six toy models (\#1-6) are divided into three groups by similar throughput. Each model  differs in the number channels  (see second column), if its  convolutions \\ are depthwise or not (see third column) and the number  of  layers in each stage.  The table shows, that ungrouped convolutions that have a higher MAC count \\ (more  than 10$\times$ the amount)  can still  be  similarly fast or faster as depthwise convolutions. Bold entries refer to the best value in each group and column
\end{tablenotes}

    \end{threeparttable}
   
\end{table*}


\section{Contrasting MACs and Latency for Vision Backbone Designs}
\label{sec:speedexperiments}

To create efficient vision backbone architectures, it is not sufficient to assess them only by the amount of their MAC operations \citep{mobileone, fasternet}. Hardware efficiency needs to be taken into account, as it is a crucial factor for the execution time of a model \citep{mobilenetv4}. For example, a model can be considered  more hardware efficient than a compared model, when it executes the same amount of MAC operations in less time.

Modern backbone architectures widely utilize convolutions as a core component \citep{fastvit,efficientnet,efficientvit,biformer}, yet their impact on hardware efficiency remains largely unexamined.
Therefore, we will investigate the hardware efficiency of convolutions in different scenarios. 
We will show that grouping (e.g. depthwise convolutions) and operating resolution of convolutions can have a substantial effect on the hardware efficiency. 
We will further analyze under which configurations (resolution and channel dimension)  it is beneficial to replace the mobile inverted bottleneck (MBConv) block \citep{mobilenetv2}, which is a popular component of many backbone architectures \citep{efficientnet, efficientvit}, with the fused MBConv \citep{fusedmbconv}.
At last, we analyze the efficiency of different adaptations of the original MHSA \citep{attentionisallyouneed} under different input resolutions.

To quantify hardware efficiency, we measured latency and image throughput across commonly used deployment platforms, including: a server machine with an Nvidia A40 GPU; a desktop machine with a Nvidia Titan RTX GPU; a GPU-capabled embedding device Nvidia Jetson TX2; an iPhone 13 smartphone with an Apple A15 Bionic; and a Raspberry Pi5 with an ARM CPU. Following previous practices \citep{shvit, efficientvit}, we run a model with a batch size of 200 for GPU throughput, while to measure latency we use a batch size of 1. We always use the median time of all iterations for latency and also to calculate throughput. As input images, we generate random tensors beforehand.

\subsection{Depthwise Convolutions have low Hardware Efficiency}
\label{subsec:grouping}

Previous research demonstrated that when optimizing backbone architectures for mobile-friendly design and efficiency, depthwise convolutions serve as a prominent alternative to standard convolutions \citep{mobilenetv3, efficientnet, mobilenetv4}. 
\bmhead{Grouped Convolutions}
Depthwise convolutions group their channels for computation. They divide them into as many groups as input channels. As a result, each feature map (channel) is  processed independently by the depthwise convolution. Therefore each output feature map is only computed by processing its corresponding input feature map.  
In contrast, standard convolutions \citep{shufflenetv2} are ungrouped, meaning all input channels contribute to every output channel (groups = 1) during the convolutional operation. 
 
\bmhead{Motivation}
Evaluating models solely based on their MACs often encourages incorporating as many depthwise convolutions as possible into the architecture, without considering their actual execution time speed-up. The efficiency of depthwise convolutions in terms of MACs does not always translate to equivalent gains in execution time.
To illustrate the discrepancy between theoretical MAC efficiency and practical execution efficiency, we conduct an experiment using a simplified toy architecture.

\bmhead{Setting}
In \Cref{tab:groupingexp}, we examine the effect of using depthwise convolutions instead of ungrouped ones on throughput and latency. We created three models with only depthwise convolutions (\#2,\#4,\#6)  and three with only ungrouped ones (\#1,\#3,\#5). Each model features five stages, mimicking common backbone architecture design. Consequently, with an input of resolution of 224×224, the output is of resolution 7×7. For the depthwise models, we doubled the amount of layers for each stage and increased the channel dimension to close up on the ungrouped models in terms of MACs. The depthwise models still have no more than a 10th of the MACs of the ungrouped convolutional models. Matching MACs -- e.g. by repeating the depthwise convolutions 20 additional times -- would result in an unusually high number of layers, deviating significantly from typical architectural designs.

\bmhead{Results}
Even though model \#1 and \#2 in \Cref{tab:groupingexp} have a similar GPU throughput, their latency differs greatly. On the Jetson TX2 and ARM CPU the ungrouped convolutions are considerably faster in most scenarios, while on desktop GPU the opposite is true. This is due to higher memory bandwidth on the desktop GPU, which is particularly important for efficient execution of depthwise convolutions \citep{mobilenetv4}. This relationships also mostly holds true for the other models (\#3-\#6). 
Regarding mobile execution, the difference in latency is not significant.
Even in the best-case scenario, depthwise convolutions require at least three times the processing time for the same number of MAC operations as their ungrouped counterparts.
In the worst scenario (\#5 and \#6), the ungrouped convolutions execute MACs ~34× faster on the Jetson TX2, highlighting the substantial hardware inefficiency of depthwise convolutions.

\bmhead{Conclusion}
In summary, ungrouped convolutions are more hardware efficient than depthwise convolutions, processing the same number of MACs in less time. However, the extent of this difference varies depending on the device.

\newenvironment{myfont}{\fontfamily{\sfdefault}\selectfont}{\par}
\begin{figure}
\centering

\includegraphics[width=\linewidth]{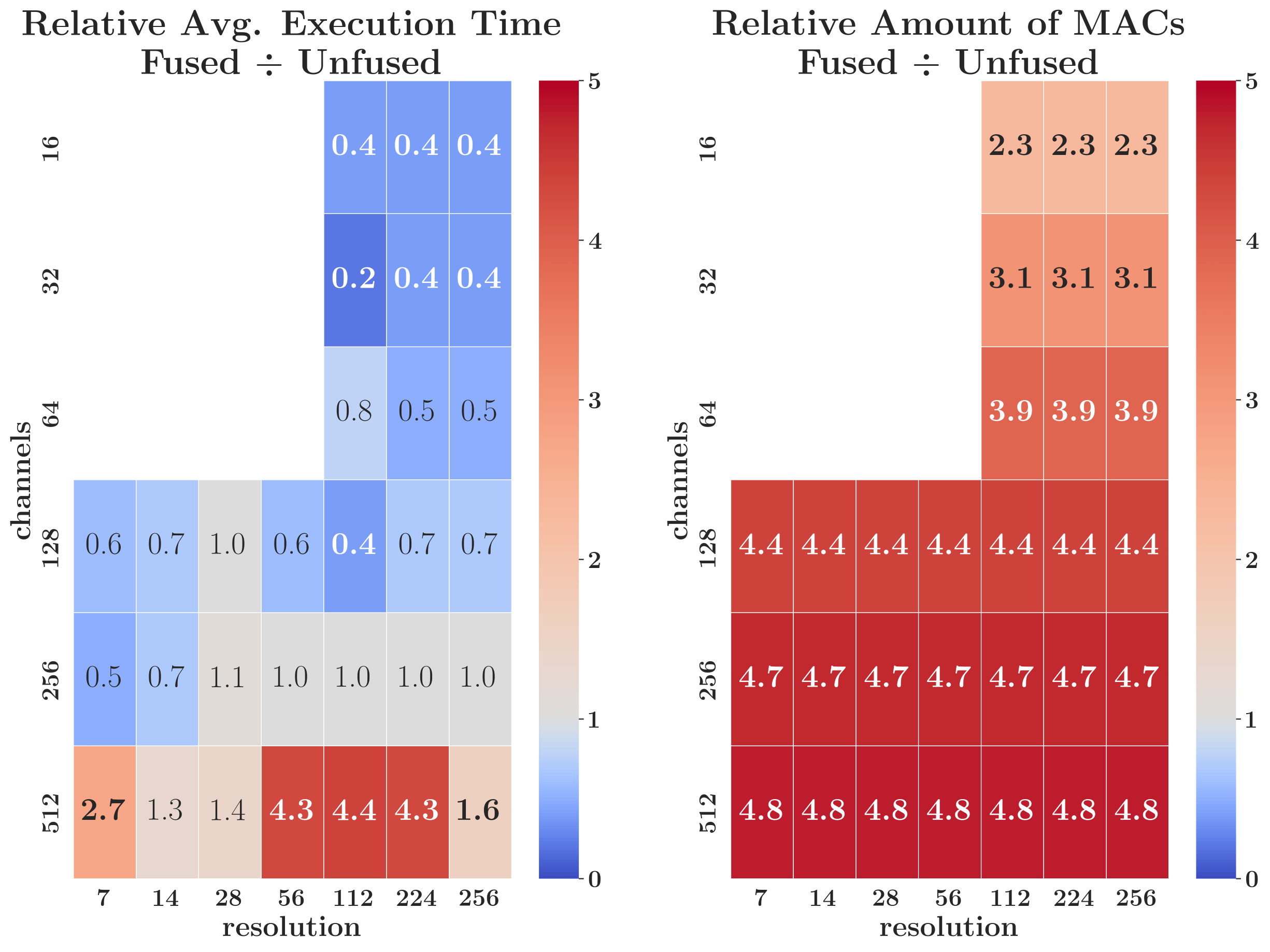}

{ }{ }{ }{ }{ } { }{ }{ }{ }{ } { }{ }{ } a)   { }{ }{ }{ }{ }{ }{ } { }{ }{ }{ }{ }{ }{ }{ }{ }{ }{ }{ }{ }{ }{ }{ }{ }{ }{ }{ }   b)

\includegraphics[width=\linewidth]{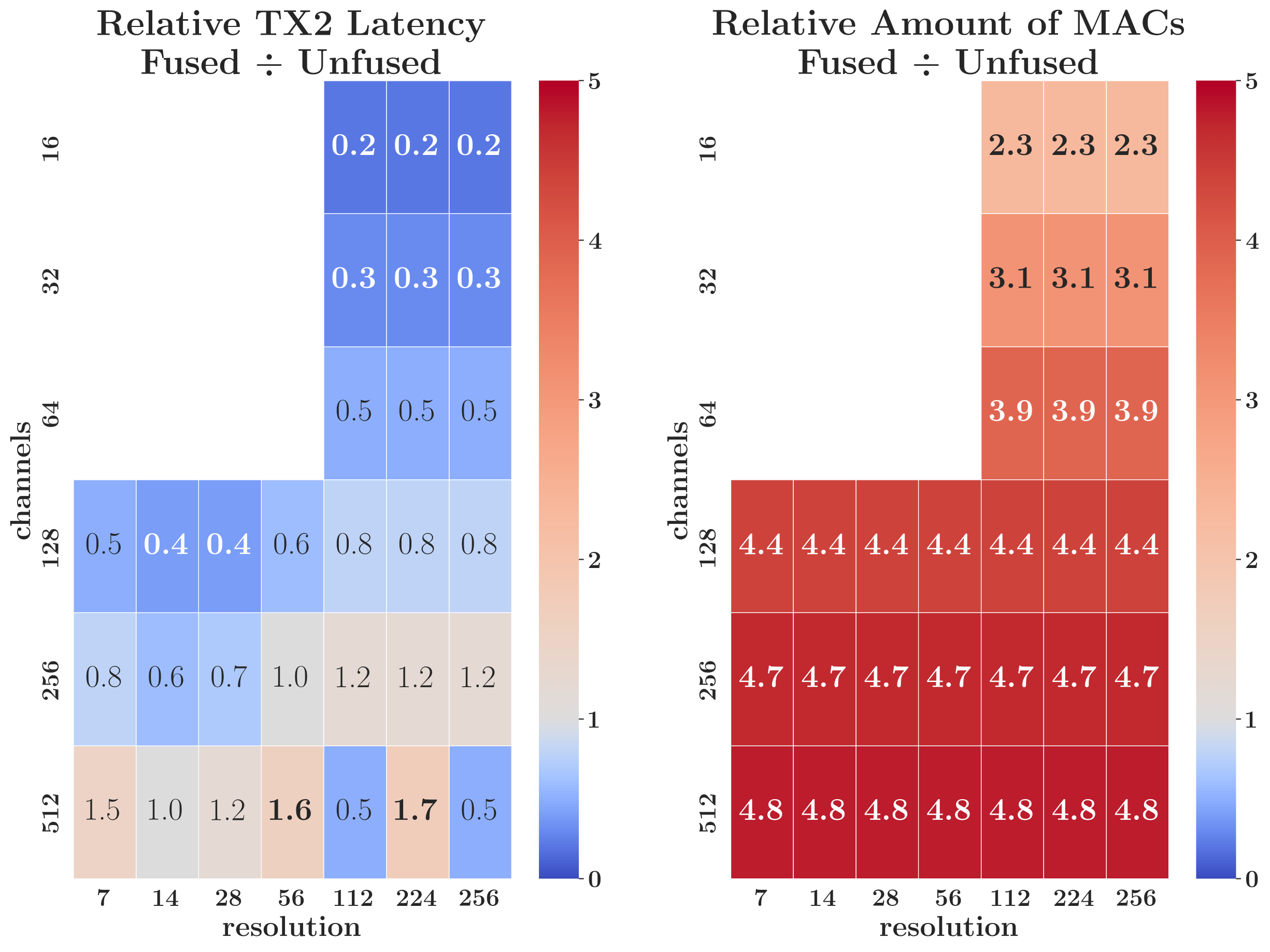}
 { }{ }{ }{ }{ } { }{ }{ }{ }{ } { }{ }{ }{ } c)  { }{ }  { }{ }{ }{ }{ }{ }{ }{ }{ }{ }{ }{ }{ }{ }{ }{ }{ }{ }{ }{ } { }{ }{ }{ }{ }  d) 
  \caption{ The left figures a) and c) depict the average execution time (a) and latency (c) of the fused mobile inverted bottleneck (MBConv)
relative to the unfused one, while the right figures b) and d) depict the relative amount of MACs.  The blue areas in the left figures a) and c) correspond to configurations (number of channels and resolution) where fused MBConv is faster, while red corresponds to the opposite. For the figures b) and d), the red areas correspond to configurations where the fused MBConv has a higher amount of MACs. Bold  numbers refer to entries with a particularly unequal ratio. Even though the fused MBConv always has more MACs, it is faster for many configurations }
  \label{fig:fusedexp}

\end{figure}

\subsection{Fusing the MBConv Block can speed up Models}

\label{subsec:fusedunfused}

The mobile inverted bottleneck block (MBConv) \citep{mobilenetv2} has a successful history in efficient backbones \citep{efficientnet,mobilenetv2,mobilenetv3} and is still used within many new architectures \citep{efficientvit,CoatNet,efficientnetv2}. It consists of two pointwise convolutions (PWConv) and a depthwise (DWConv) in between (see \Cref{fig:mbconv} for a depiction of it). The PWConvs increase and decrease the channel dimension by the attributed expansion factor. 
An alternative to the MBConv block is the fused MBConv block \citep{fusedmbconv}. The latter merges the first PWConv with the DWConv into an ungrouped convolution, thus removing any depthwise convolution.
Due to the depthwise convolution, the original MBConv block usually has a lower amount of MACs than the fused one, although it features one layer more.

\bmhead{Motivation}
In \Cref{subsec:grouping}, we showed that depthwise convolutions are hardware inefficient. Here, we extend that experiment to obtain  insights into efficient backbone macro-design. We measure under which operating resolution and channel dimension the fused MBConv block (Conv+PWConv) is faster than the original MBConv  (PWConv+DWConv+PWConv), allowing us to determine the optimal choice for different parts of the architecture.

\bmhead{Setting}
For this experiment, we measure latency on the Nvidia Jetson TX2 and average execution time on a Nvidia A40 (see \Cref{fig:fusedexp}). 
To calculate the average execution time, we run 100 iterations with a batch size of 200, take the median execution time across these iterations, and divide it by 200.
This is the inverse of the throughput metric and ensures consistency in the presentation of the sub-figures in \Cref{fig:fusedexp}.

We apply an expansion factor of 4 for the fused and original MBConv in \Cref{fig:fusedexp}.
For this, we created toy models that just consist out of the same layer repeated after another.

\bmhead{Relative execution time}
In \Cref{fig:fusedexp}, each metric is a relative metric, meaning it is always the value of the fused MBConv divided by the unfused one.
For example in a) of \Cref{fig:fusedexp}, we depict  the relative average execution time, meaning the average execution time of the fused MBConv divided by the unfused one. In b) and d), we depict their relative MAC count (fused divided by unfused) and in c) their relative latency on the TX2. 
We apply these metrics for various resolutions and input channel dimensions. We omit the scenarios where channel dimension range from 16 to 64 and resolutions from 7 to 28, as GPU utilization is too low for the results to have significance.

\bmhead{Results}
It can be noticed that resolution and channel dimension have a big influence on the relative latency c) of \Cref{fig:fusedexp} and average execution time a). Even though the fused MBConv always has more MACs --values over one in right part b) and d) --, it is faster in many scenarios -- i.e., values smaller than one in a) and c) -- and is always more hardware efficient, i.e. the relative latency/average execution time is smaller than relative MAC count.
Only for an high number of channels ($>=$256) and an high operating resolution ($>=$56), the fused MBConv presents slower execution. 
For channel dimension 512, and  resolutions 112 and 256, the fused MBConv is faster than the unfused one. However, for resolution 224, it is the other way around. 
This inconsistency occurs because the computational load and memory requirements for these configurations exceed the Jetson TX2's capacity, leading to unpredictable behavior.

\bmhead{Conclusion}
The efficiency of the fused MBConv heavily depends on the channel dimension and to a lesser extent on the operating resolution. Consequently, it is advantageous to apply the fused MBConv in the early stages of an architecture, where the channel dimension is typically low ($<$256). In later stages that feature a higher channel dimension, the original MBConv proves more efficient.

\begin{table*}[t]
    \centering

    \begin{adjustbox}{max width=0.9\textwidth,center}
    
       \begin{threeparttable}[b]
       \caption{ \fontsize{9}{10}\selectfont  Experiment on the impact of resolution on hardware efficiency }
    \label{tab:highreshighchan}
    
   \begin{tabular}{c|c|c|c?c|c|c|c} 
    \Xhline{1.5pt}
        
         \multirow{2}{*}{Scenario} & Resolution  & Channels & Relative & Relative Throughput $\uparrow$ & \multicolumn{3}{c}{Relative Latency (ms) $\downarrow$}   \\
        
          & (pixel) & (\#) & MACs & GPU (images/s) & \multicolumn{1}{c}{TX2} & \multicolumn{1}{c}{Mobile} & \multicolumn{1}{c}{ARM CPU}  \\ 
          \Xhline{1.5pt}

        \multirow{2}{*}{\#1}  & 224 & 24 & 1.0 &  0.3 & 1.71 & 1.26 & 1.42   \\ 
        
       {} & \cellcolor{lightergray}  28 &  \cellcolor{lightergray} 196  & \cellcolor{lightergray}  1.0 &  \cellcolor{lightergray}  \textbf{3.3}& \cellcolor{lightergray} \textbf{0.58} & \cellcolor{lightergray}  \textbf{0.79} & \cellcolor{lightergray}  \textbf{0.70}     \\  
       \Xhline{1.5pt}

        \multirow{2}{*}{\#2}  & 224 & 48  & 1.0 & 0.5 & 1.60 & 1.22 & 1.16   \\   
       & \cellcolor{lightergray}   112 & \cellcolor{lightergray} 96  & \cellcolor{lightergray}  1.0 & \cellcolor{lightergray}   \textbf{1.9} & \cellcolor{lightergray}  \textbf{0.63 } & \cellcolor{lightergray}  \textbf{0.82} & \cellcolor{lightergray}  \textbf{0.86}   \\  
       \Xhline{1.5pt}
        

        \multirow{2}{*}{\#3}  & 224 & 96  & 1.0 & 0.5 & 1.06 & 1.02 & 1.09   \\  
       & \cellcolor{lightergray}  56 & \cellcolor{lightergray}  384  & \cellcolor{lightergray}  1.0 &  \cellcolor{lightergray}   \textbf{1.9} & \cellcolor{lightergray} \textbf{0.94} & \cellcolor{lightergray}  \textbf{0.98}  & \cellcolor{lightergray}  \textbf{0.92}    \\  
       \Xhline{1.5pt}
        

           \multirow{2}{*}{\#4}  & 224 & 48 & 1.0 & 0.5 & 1.16 &  1.17 & 1.02     \\  
         & \cellcolor{lightergray}  56 & \cellcolor{lightergray}  196  & \cellcolor{lightergray}  1.0 & \cellcolor{lightergray}    \textbf{2.0} & \cellcolor{lightergray}  \textbf{0.86} & \cellcolor{lightergray}  \textbf{0.85} & \cellcolor{lightergray}  \textbf{0.98}    \\    
         \Xhline{1.5pt}
        
        \multirow{2}{*}{\#5}  & 112 & 24 & 1.0 & 0.3 & 1.62 & \textbf{0.89} & 1.12    \\  
       & \cellcolor{lightergray}  14 & \cellcolor{lightergray}  196  & \cellcolor{lightergray}  1.0 & \cellcolor{lightergray}    \textbf{3.3} & \cellcolor{lightergray}  \textbf{0.62} & \cellcolor{lightergray}   1.12 & \cellcolor{lightergray}   \textbf{0.89}   \\   
       \Xhline{1.5pt}

    \multirow{2}{*}{\#6}  & 56 & 96  & 1.0 & 0.5 & \textbf{1.0} & \textbf{0.78} & \textbf{0.74} \\  
       & \cellcolor{lightergray}   14 & \cellcolor{lightergray}  384  & \cellcolor{lightergray}  1.0  & \cellcolor{lightergray}   \textbf{2.2}& \cellcolor{lightergray}  \textbf{1.0} & \cellcolor{lightergray}  1.28 & \cellcolor{lightergray}  1.35   \\ 
       \Xhline{1.5pt}
        
        \multirow{2}{*}{\#7}  & 112 & 96 & 1.0  & 0.6 & \textbf{0.88} & \textbf{0.85} & \textbf{0.99}   \\ 
        & \cellcolor{lightergray}  28 & \cellcolor{lightergray}  384 & \cellcolor{lightergray}  1.0  & \cellcolor{lightergray}    \textbf{1.8} & \cellcolor{lightergray}  1.14 & \cellcolor{lightergray}  1.18 & \cellcolor{lightergray}  1.01   \\  
        \Xhline{1.5pt}

    \end{tabular}
  \begin{tablenotes}[para,flushleft]
        \fontsize{8}{9}\selectfont
      Each scenario contains two configurations of
convolutions (first row and second row in each scenario), that approximately feature  the same amount of MACs (see 4th
column), but strongly differ in operating resolution and number of channels. We set their  throughput and latency in
relation to each other(see 5th-8th column). The table shows, that convolutions operating on a  higher  resolution tend to be
less efficient than convolutions operating on a lower resolution (highlighted in gray). Bold values refer to the best value for each scenario and column
  \end{tablenotes}
  
  \end{threeparttable}
 \end{adjustbox}

\end{table*}

\subsection{High Resolution vs. High Channel}
\label{subsec:highreshighchan}

\bmhead{Motivation}
A key aspect of architecture design is determining the distribution of layers across different stages of the model. In backbone architectures, early stages typically have a high operating resolution and a low channel dimension, whereas later stages the opposite trend \citep{efficientnet, efficientvit,mobilenetv4}. Therefore, understanding whether layers with a high channel dimension and low operating resolution are more efficient than those with a low channel dimension and high operating resolution is crucial for efficient architecture design. This insight helps optimize the allocation of layers across stages.
In the following, we will analyze this factor for convolutional layers.

\bmhead{Setting}
For this experiment, we create toy models  that consist of 20 times the same convolution stacked after another, where each toy model has a different configuration regarding operating resolution and channel dimension.
In \Cref{tab:highreshighchan}, we put at test seven scenarios, each contrasting two convolutions with the same amount of MACs, but differing in channel dimension and operating resolution. 
The models only contain standard convolutions (ungrouped).
The upper row in each scenario is always the model operating on a higher resolution, while the lower row features a higher channel dimension. 
For improved readability we state relative throughput, latency and MAC count in \Cref{tab:highreshighchan}. In each scenario, the metrics (throughput, latency, MAC count) of the low resolution model are expressed relative to those of the high resolution model and vice versa for the high resolution model.

\bmhead{Results}
In scenario \#1 the first model has a third of the throughput of the second one and almost twice the latency. It operates on eight times the resolution, however features less channels and its MACs equal the second layer.
The same effect also occurs with a smaller resolution difference, as scenario \#2 shows, where the first model runs on twice the resolution, but still fails to execute its MACs as fast as the second one in terms of throughput and latency.
On the other side in scenario \#7 the model with a higher operating resolution, has a slightly lower latency. Regarding GPU throughput however, the lower resolution models always have a considerably higher throughput, ranging from a factor of 1.8 to 3.3. 
In scenario \#5 the lower resolution model achieves lower latencies for the TX2 and ARM CPU, while for mobile the higher resolution model prevails, showing the impact that different hardware can have on model execution. 
Nevertheless a clear trend is visible. Models with a high operating resolution tend to be slower in most scenarios in terms of latency and throughput, than their MAC count might suggest.
In \citep{nottebaum2024lowformer} this is also shown for GPU latency.

\begin{table*}[ht]
    \centering
        \caption{Efficiency comparison between original MHSA and three adaptations of it } 
    \label{tab:attentionspeedexp}

           \begin{threeparttable}[b]
       
    \resizebox{1.0\textwidth}{!}{\begin{tabular}{c|c|c|c|c|c||c}
        \Xhline{1.5pt}
    
         & \diagbox{Attention}{Resolution} & 8×8 & 16×16 & 32×32 & 64×64 & avg. diff. \\
            \Xhline{1.5pt}
            &  \cellcolor{lightergray}   MHSA & \cellcolor{lightergray}  0.59 & \cellcolor{lightergray}  2.79 & \cellcolor{lightergray}  19.70 & \cellcolor{lightergray}  260.97  & \cellcolor{lightergray}  \\
        
          TX2 Latency   & chcompr. & 0.38 \textbf{(-37\%)} & 1.63 \textbf{(-41\%)} & 10.29 \textbf{(-47\%)} & 130.73 \textbf{(-50\%)} &  \textbf{-43\%} \\
          
            (ms) & conv+low & 0.53 \textbf{(-10\%)} & 1.39 \textbf{(-50\%)} & 4.87 \textbf{(-75\%)} & 28.09 \textbf{(-89\%)} &  \textbf{-56\%} \\

            & conv+low+chcompr. & 0.39 \textbf{(-35\%)} & 1.21 \textbf{(-56\%)} & 3.62 \textbf{(-81\%)} & 18.56 \textbf{(-92\%)} &  \textbf{-66\%} \\
            
            \Xhline{1.5pt}
            
            &  \cellcolor{lightergray}  MHSA & \cellcolor{lightergray} 0.20 & \cellcolor{lightergray} 1.60 & \cellcolor{lightergray}  5.48 & \cellcolor{lightergray}  179.75  & \cellcolor{lightergray}  \\

           Mobile Latency  & chcompr. & 0.15 \textbf{(-25\%)} & 0.79 \textbf{(-50\%)} & 2.62 \textbf{(-52\%)} & 65.10 \textbf{(-63\%)} &   \textbf{-47\%} \\
            
            (ms) & conv+low & 0.24 \textbf{(+20\%)} & 0.45 \textbf{(-71\%)} & 1.34 \textbf{(-75\%)} & 5.60 \textbf{(-97\%)} & \textbf{-56\%} \\
            
            & conv+low+chcompr. & 0.21 \textbf{(+5\%)} & 0.33 \textbf{(-79\%)} & 0.71 \textbf{(-87\%)}  & 2.71 \textbf{(-98\%)} &  \textbf{-65\%} \\
               \Xhline{1.5pt}

           & \cellcolor{lightergray}   MHSA &  \cellcolor{lightergray} 0.81 & \cellcolor{lightergray} 6.16 & \cellcolor{lightergray} 95.32 & \cellcolor{lightergray} 1808.16  & \cellcolor{lightergray}  \\
          ARM CPU Latency & chcompr. & 0.57 \textbf{(-30\%)} & 2.42 \textbf{(-60\%)} & 46.73 \textbf{(-50\%)} & 915.41 \textbf{(-49\%)} &  \textbf{-47\%}\\
            (ms) & conv+low & 0.85 \textbf{(+5\%)} & 2.21  \textbf{(-64\%)} & 10.97  \textbf{(-88\%)} & 126.48 \textbf{(-93\%)} & \textbf{-60\%} \\
            & conv+low+chcompr. & 0.75  \textbf{(-7\%)} & 1.88 \textbf{(-69\%)} & 7.23  \textbf{(-92\%)} & 75.79 \textbf{(-95\%)} &  \textbf{-66\%} \\  
        \Xhline{1.5pt}

            & \cellcolor{lightergray}   MHSA & \cellcolor{lightergray}  0.49 & \cellcolor{lightergray}  0.49 & \cellcolor{lightergray} 1.24 &  \cellcolor{lightergray} 13.85  &  \cellcolor{lightergray} \\
          
          GPU Latency & chcompr. & 0.49 \textbf{(-0\%)} & 0.50 \textbf{(+2\%)} & 0.85 \textbf{(-31\%)} & 7.76  \textbf{(-44\%)}& \textbf{-18\%} \\
          
            (ms) & conv+low & 0.62 \textbf{(+26\%)} & 0.63 \textbf{(+28\%)} & 0.63 \textbf{(-49\%)} & 1.42 \textbf{(-90\%)} & \textbf{-22\%}  \\
            
            & conv+low+chcompr. & 0.63 \textbf{(+28\%)} & 0.64 \textbf{(+30\%)} & 0.64 \textbf{(-48\%)} & 1.02 \textbf{(-92\%)}  & \textbf{-20\%}\\

        \Xhline{1.5pt}

    \end{tabular}}
    \begin{tablenotes}[para,flushleft]
    \footnotesize
        "conv+low" refers to encapsulating MHSA with convolutions that reduce the resolution of the attention by a factor of 2,  
         "chcompr." refers to a reduced channel \\ dimension for SDA and "conv+low+chcompr." refers to the combination of the latter two.  We evaluate execution time under  different  input  resolutions. \\ The highest efficiency is achieved by applying the "conv+low+chcompr." adaptation on MHSA. Results for original MHSA are highlighted in gray
    \end{tablenotes}
   \end{threeparttable}

\end{table*}

\bmhead{Conclusion}
We can conclude that it in most cases it proves more hardware efficient to apply more layers in later stages of the backbone architecture and have only few layers in the high resolution stages. 
Based on these experiments, we emphasize that model scaling (increasing width, depth and input resolution \citep{scalingvits}) should be guided by actual measured execution time \citep{efficientnet}, as MACs can be misleading, when scaling models by higher input resolution \citep{shvit,efficientvit}, as our results indicate.

\subsection{Optimizing MHSA Efficiency for Higher Input Resolutions}
\label{subsec:attentionspeedexp_chapter}

The MHSA introduced by \citep{attentionisallyouneed} is a crucial building block in many recent computer vision frameworks \citep{segmentanything, Stark,Detr,dunnhofer2022combiningattention,khan2024idenetattention,bansal2022spatioattention}. The actual attention operation takes place within the SDA module of MHSA.
\bmhead{Motivation}
A key limitation of MHSA is that the computational complexity of SDA scales quadratically with spatial dimensions and linearly with the number of channels.
To address this issue, we propose adaptations that reduce the input dimensions for SDA, enhancing efficiency without sacrificing performance. We evaluate the efficiency of both the original MHSA and our adapted versions across input resolutions ranging from 8×8 to 64×64.
This is motivated from the fact, that many frameworks incorporating MHSA typically feature a high input resolution ($>$512×512) \citep{segmentanything,  poseestimationSurvey, vityoloinputresol, Detr}, consequently leading to SDA being executed at resolutions above 16×16. This is particularly true when the backbone itself includes MHSA. For instance, in the ViT-B/16 vision backbone \citep{ViT}, an input resolution of 1024×1024 \citep{segmentanything} leads to SDA being computed at resolution 64×64.

\bmhead{Setting}
To conduct this experiment, we employ toy models that consist of 4 times the same attention layer after another. We feature an input and output channel dimension of 128 to prevent overloading the edge devices, as MHSA at increased input resolutions imposes a high computational burden.
We measure latency on the Jetson TX2, the  iPhone 13, the ARM CPU of the Raspberry Pi5 and the Nvidia TITAN RTX GPU.

\bmhead{Efficient Adaptations}
In total, we feature two modifications of MHSA: "chcompr." (channel compression) stands for halving the channel dimension before the SDA by the input projection and restoring the input channel dimension by the output projection of the MHSA. "conv+low" means we add a convolution before the input projection and after the output projection, which downsample and upsample the resolution of the feature maps.
In \Cref{tab:attentionspeedexp}, we compare how these two modifications, taken together and separately, influence the execution time under different input resolution (from 8×8 to 64×64).

\bmhead{Results}
In \Cref{tab:attentionspeedexp} we compare MHSA with the two adaptations we propose.
\Cref{tab:attentionspeedexp} illustrates the impact of the quadratic explosion of MHSA. On edge devices, latency is at least 442 times higher for a resolution of 64×64 compared to 8×8, and even 2232 times higher on an ARM CPU. In contrast, GPU execution shows only a 28-fold increase in latency, highlighting the immense parallelization capabilities of GPUs compared to edge devices.
Regarding our adaptations, we can see that "chcompr." leads to a reduction in latency of between 30\% and 63\% on edge devices, while "conv+low" leads to an even higher efficiency gain. The performance improvement for both adaptations increases significantly with higher input resolution. On GPU with input resolutions 8×8 and 16×16, the efficiency remains similar to MHSA  due to the potential of parallelization a desktop GPU has  or slightly declines, because of the layers added by "conv+low" adaptation. Nevertheless for input resolutions above 16×16, both adaptations lead to a considerable reduction in latency by up to 90\% compared to MHSA. Combining both methods ("conv+low+chcompr") maximizes efficiency. For input resolution 64×64 its latency reduction ranges between 92\% and 98\% across all devices.

\bmhead{Conclusion}
While "chcompr." can lead to a reduction of up to half of the latency, "conv+low" can reduce latency to a smaller fraction of it.
Overall, the combination of the two optimizations ("chcompr." and "conv+low") provides significant benefits for performance, particularly for edge devices.
At high input resolutions, the adaptations also lead to substantial performance gains for GPUs. 
Since many downstream tasks rely on high input resolution, these optimizations are highly relevant \citep{segmentanything,Detr,groundingdino}.
In \Cref{subsec:ablation}, we will further demonstrate that the combination of both optimizations has a positive impact on ImageNet \citep{imagenet} accuracy.

\begin{figure*}[hbt!]
  \centering
  
    \begin{minipage}[b]{0.25\textwidth}
    \includegraphics[width=0.85\textwidth]{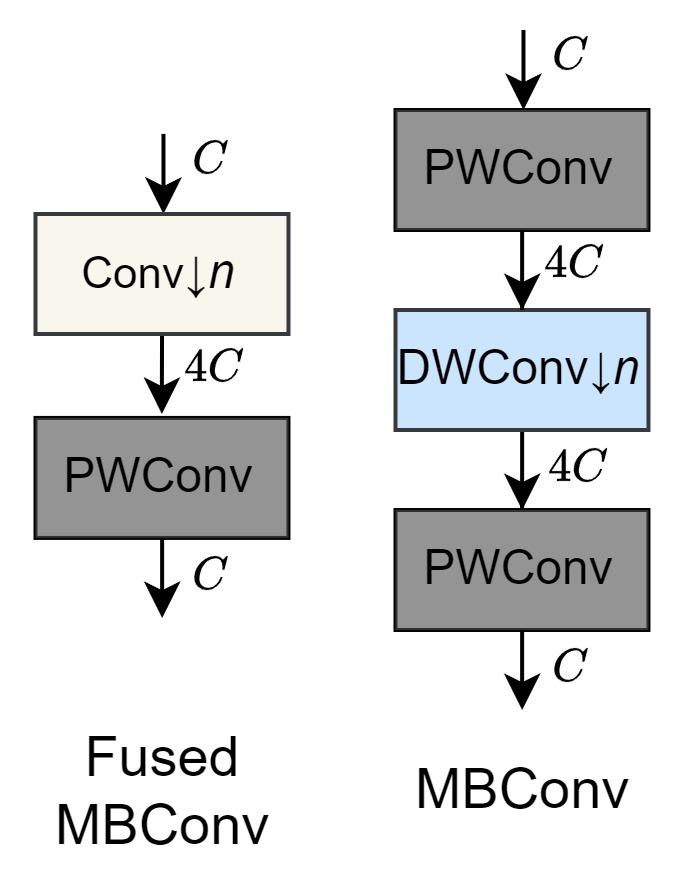}
    \caption{Structure of the fused and unfused MBConv block. $C$ refers to the channel dimension. Both have an expansion factor of 4} 
    \label{fig:mbconv}

  \end{minipage} \hfill \begin{minipage}[b]{0.7\textwidth}

 \includegraphics[width=\textwidth]{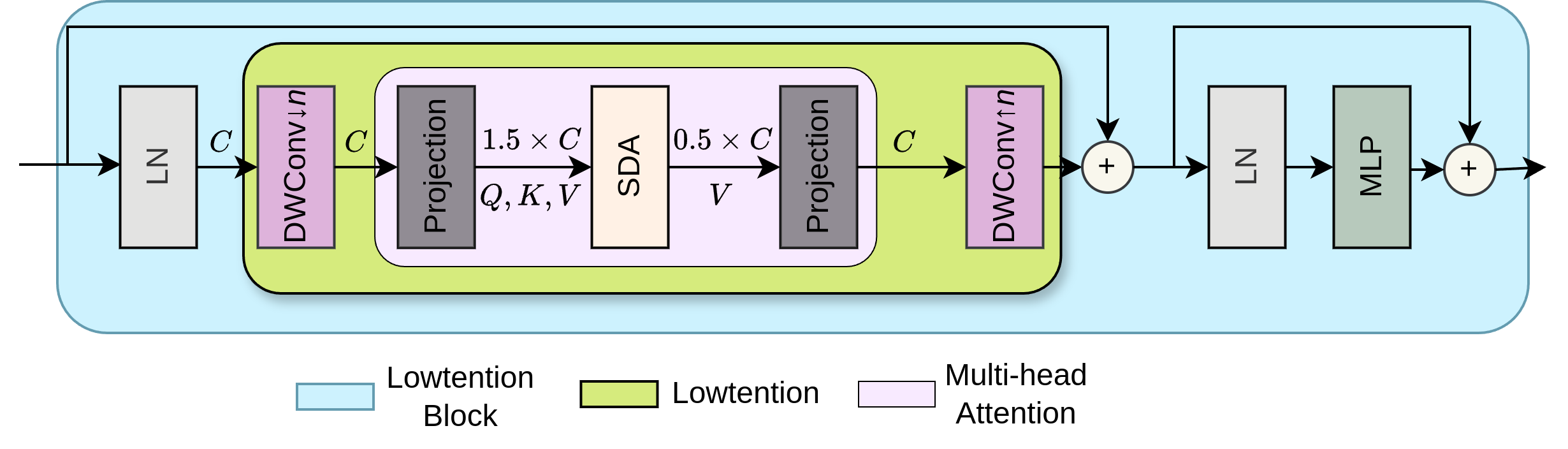}

         \caption{\lowformeratt { }block design. LN refers to layer normalization. 
  In contrast to the traditional MHSA, we encapsulate the SDA with two depthwise convolutions (the second is a transposed depthwise convolution). The projections for MHSA are realized with pointwise convolutions. The  $DW\downarrow_n$ means that the resolution is downscaled by the factor $n$ and $DW\uparrow_n$ that it is upscaled by $n$}
  \label{fig:lowformerblock}

  \end{minipage} 
\end{figure*}

\begin{figure*}[hbt!]

     \includegraphics[width=\linewidth]{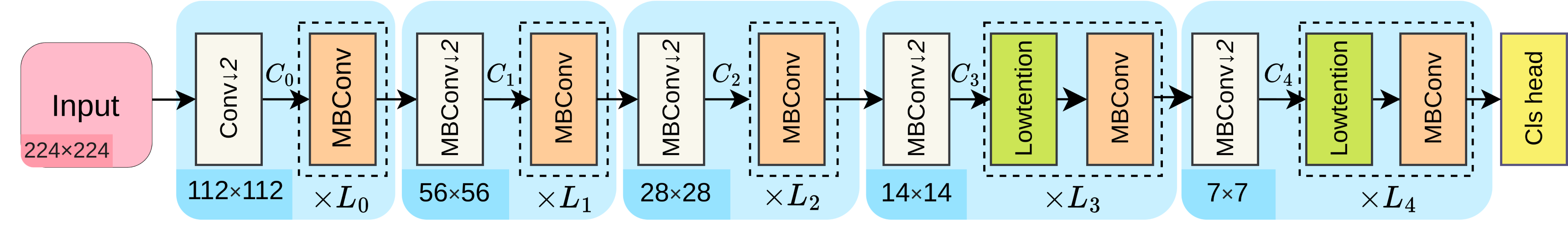}
  \caption{Architecture of LowFormer. The resolutions 
refer to a 224x224 sized input. \lowformeratt { }block can be seen in  \Cref{fig:lowformerblock}. "Conv" refers to convolution and "Cls head" refers to the image classification head (see \Cref{fig:arch_class_head}). Specification of $C_0 - C_4$ and $L_0 - L_4$ depend of the version of LowFormer and can be found in \Cref{tab:archnumbers} }
  \label{fig:arch}

\end{figure*}


\section{LowFormer}
\label{sec:method}

The aim of the LowFormer architecture design is to improve hardware efficiency, allowing vision backbones to execute faster while maintaining a high 
model size
that enables reaching high accuracy.
To achieve this goal, we follow the insights of the analysis presented in  \Cref{sec:speedexperiments}.  
In this section, we will first introduce \lowformeratt, a lightweight adaptation of the original 
MHSA \citep{attentionisallyouneed} (\Cref{subsec:microdesign}),
and then outline the key principles of LowFormer's macro design and provide both an overview (\Cref{subsec:macrodesign}) and detailed description of the overall architecture (\Cref{subsec:architecture_details}).
Finally, we will propose adaptations to the main LowFormer architecture aimed specifically at further enhancing performance on edge devices (\Cref{subsec:adapting_lowformer_for_edge}).

\subsection{Micro Design - \lowformeratt}
\label{subsec:microdesign}
In LowFormer's architecture, we employ a lightweight adaptation of the original MHSA \citep{attentionisallyouneed}, which we call \lowformeratt{ }  (see \Cref{fig:lowformerblock}). 
\lowformeratt { } encapsulates the SDA
by two depthwise convolutions and two pointwise convolutions. The latter perform the input and output projections of the queries (Q), keys (K) and values (V), just as in the original MHSA. 
\bmhead{Channel compression}
However, in \lowformeratt { }the input projection (before the SDA) reduces the channel dimension of Q, K and V by half (see \Cref{fig:lowformerblock}). 
The output projection (following the SDA) then restores the original channel dimension, which  is required for the residual connection after the \lowformeratt { }(see \Cref{fig:lowformerblock}).
In \Cref{subsec:attentionspeedexp_chapter}, we demonstrated that compressing the channel dimension can significantly improve latency on a variety of devices, especially at high input resolutions.
\bmhead{Lower resolution}
As depicted in \Cref{fig:lowformerblock}  we down- and upsample the resolution of the feature maps in \lowformeratt { }around the SDA, such that the SDA is executed on half the resolution. This motivates from the experiments in \Cref{subsec:attentionspeedexp_chapter}, where we have shown that this can improve latency significantly, similar to the channel compression in \lowformeratt.

In \Cref{subsec:ablation} we further show that the combination of both strategies (channel compression and lower resolution) improves top-1 accuracy on ImageNet \citep{imagenet}.

\bmhead{MLP following Attention}
Following \citep{attentionisallyouneed}, we append layer normalization and a multi-layer perceptron (MLP) after the \lowformeratt. 
This is motivated by \cite{efficientvitmemory}, who pointed out the significance of MLPs for improving accuracy of a backbone.

\subsection{Macro Design}
\label{subsec:macrodesign}
The LowFormer architecture features five stages that adapt the architectural macro design of EfficientViT \citep{efficientvit} and MobileViT \citep{mobilevit} according to the insights gained from the hardware efficiency analysis presented in \Cref{sec:speedexperiments}. 
The whole architecture is depicted in \Cref{fig:arch}. 

In total we present five different base versions of LowFormer, namely B0, B1, B1.5, B2, B3.
The versions differ in the number of layers and channel dimension employed in each stage (see \Cref{tab:archnumbers}).
We chose to feature five base versions of LowFormer to demonstrate that its design principles are adaptable to various model sizes and accuracy levels.
LowFormer-B0 represents the model with the lowest model size, while LowFormer-B3 has the highest. Consequently a base version with a higher model size also achieves superior results compared to a lower size variant, as shown in  \Cref{sec:Experiments}.

\begin{table}[bt!]
    \fontsize{8}{9}\selectfont
\setlength\tabcolsep{.14cm}
    \centering
       \caption{Specification of LowFormer architecture versions B0-B3  }
    \label{tab:archnumbers}

    \begin{tabular}{c|c|c}
        \Xhline{1.0pt}
         Model  & $\{L_0,L_1,L_2,L_3,L_4 \}$  & $\{C_0,C_1,C_2,C_3,C_4 \}$  \\
         \Xhline{0.5pt}

         LowFormer-B0 & $\{0,0,0,3,4\}$ & $\{16,32,64,128,256\}$  \\
         LowFormer-B1 & $\{0,0,0,5,5\}$ & $\{16,32,64,128,256\}$  \\
        LowFormer-B1.5 & $\{0,0,0,6,6\}$ & $\{20,40,80,160,320\}$  \\
         
         LowFormer-B2 & $\{0,0,0,6,6\}$ & $\{24,48,96,192,384\}$  \\
         LowFormer-B3 & $\{1,1,2,6,6\}$ & $\{32,64,128,256,512\}$  \\
        
        \Xhline{1.0pt}
    \end{tabular}
    
     \begin{tablenotes}[para,flushleft]
        \footnotesize
        The number of layers ($L_0-L_4$) and channels ($C_0-C_4$) relates to \Cref{fig:arch}
   \end{tablenotes}
 
\end{table}

\bmhead{\lowformeratt}
We include \lowformeratt { }in the last two stages and keep the first three stages purely convolutional. Additionally, we only downsample the feature maps of \lowformeratt { }(as mentioned in \Cref{subsec:microdesign}) in the forelast stage.

\bmhead{Fusing Depthwise and Pointwise Convolutions}
In \Cref{subsec:grouping} we showed that depthwise convolutions are not as hardware-efficient as standard convolution and in \Cref{subsec:fusedunfused} we came to the conclusion that the fused MBConv (see \Cref{fig:mbconv}) can be faster than the unfused one, even though it usually has a higher MAC count. This effect diminishes however with increasing number of channels. We therefore fused the MBConv in our architecture, wherever the number of input channels reach at most 256, except for the strided MBConv blocks at the beginning of the last two stages (see \Cref{fig:arch}).
We additionally fuse the depthwise and pointwise convolutions after the SDA in the \lowformeratt{ } (see \Cref{fig:lowformerblock}), as their input channel dimension does not exceed 256 for any LowFormer model, due to the channel compression. 
We confirm the effect of this approach by reverting the fusion of the MBConv block for LowFormer-B1 in the ablation in \Cref{subsec:ablation}.

\bmhead{Less layers in the first stages}
From the insights in \Cref{subsec:highreshighchan} we conclude that a minimal amount of layers in the first stages is more hardware-efficient (see \Cref{tab:archnumbers}).
It proved optimal to apply the reduction for the first three stages. Most computation is therefore concentrated in the last two stages, where for an input size of 224$\times$224, the operating resolutions are 14$\times$14 and 7$\times$7.

\subsection{Additional Architectural Details}
\label{subsec:architecture_details}

The details of LowFormer's architecture mainly follow design principles of previous publications \citep{efficientvit,fastvit}.

As activation function, we utilize HardSwish \citep{mobilenetv3}, except for the MLP, where we use GeLU \citep{geluactfunc} in between its two linear layers \citep{attentionisallyouneed}.

\bmhead{MBConv Micro Design Details}
LowFormer's micro design for MBConv blocks, regarding the combination of batch normaliation (bn) \citep{batchnormpaper} and activation functions (act), differs between fused and original (unfused) MBConv. 
For the fused MBConv, each block follows the design of "conv,bn,act,pwconv,bn", while the original MBConvs design is "pwconv,act,dwconv,act,pwconv,norm".

\bmhead{MBConv Expansion Factor}
The expansion factor in MBConvs controls by which factor the channel dimension is increased by the first convolution and decreased by the last convolution (see \Cref{fig:mbconv}). We set it to 6 for all MBConv blocks (fused and unfused) that reduce resolution and 4 otherwise. 
\bmhead{Residual Connections}
Besides the residual connections in the \lowformeratt{ } block (see \Cref{fig:lowformerblock}), every MBConv block that does not reduce resolution, has a residual connection. We only reduce resolution by strided MBConv blocks at the beginning of each stage.

\begin{figure}[t]
 
   \includegraphics[width=\linewidth]{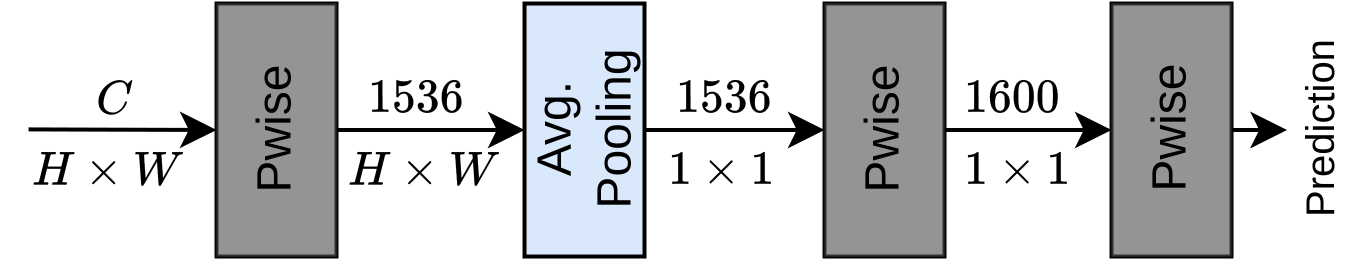}
  \caption{Design of classification head that is used for all variants of LowFormer}
  \label{fig:arch_class_head}

\end{figure}

\bmhead{Classification Head}
In \Cref{fig:arch_class_head} the design of our classification head is depicted, which we apply for all versions of LowFormer and only differs in the input channel dimension, given by the last layer of the final stage. It is based on EfficientViTs \citep{efficientvit} classification head.

\subsection{LowFormer for Edge GPU}
\label{subsec:adapting_lowformer_for_edge}
We further present three edge GPU variants of LowFormer, namely LowFormer-E1/E2/E3.
They are derivations of the so far described architecture. The changes only focus on the last two stages, as most of the computational load is concentrated there (see \Cref{tab:archnumbers}). 
We feature three changes to the original LowFormer Architecture: 
\begin{enumerate}[label=\alph*)]
    \item Reduction of the depth by 2
    \item Removal of the MVLP
    \item Removal of the \lowformeratt
\end{enumerate}
All changes follow the intuition that Edge GPUs (like the Jetson TX2) usually consist of  many  cores\footnote{\url{https://developer.nvidia.com/embedded/jetson-tx2}} and focus on high parallelization, similar to GPUs\footnote{\url{https://www.nvidia.com/content/dam/en-zz/Solutions/titan/documents/titan-rtx-for-creators-us-nvidia-1011126-r6-web.pdf}}. This is in contrast to CPUs for example, who usually feature much less cores, like the ARM CPU in the Raspberry Pi5, who operates on 4 cores\footnote{\url{https://www.raspberrypi.com/products/raspberry-pi-5/}}.

Change a) therefore originates from the intuition, that depth is a crucial factor in optimizing efficiency on edge GPU devices, as the compulsory sequential execution of high amount of layers, prohibits parallelization, thus it would be beneficial to rather have fewer layers, but with high amount of computation, that is parallelizable. 

Change b) motivates from the observation of \citep{mobilenetv4}, that MLP executions are usually memory bound and therefore are not efficient on hardware with high compute ability (e.g. GPUs and edge GPUs). Additionally MLPs account for a significant portion of the whole models computational load (MACs). For LowFormer-B1 for example, the MLPs make up 17\% of the total MACs. 

Change c) on the other side motivates from our observation  in \Cref{subsec:attentionspeedexp_chapter}, where the attention mechanisms show an immense computational burden, especially with increased input resolution. Furthermore,  \citep{efficientvitmemory} demonstrated that MHSA operations are heavily memory bound, even more so than the MLP. They recommend allocating a lower portion of the model to MHSA to yield the best speed accuracy trade-off.

We combine the three changes a), b) and c) into the models LowFormer-E1/E2/E3 as depicted in \Cref{tab:edgevariant_archnumbers}. 
In \Cref{subsec:eff_analysis_edgebones}, we substantiate our reasoning  by comparing how each mentioned change impacts accuracy and efficiency for edge GPU and GPU.

\begin{table}[bt!]
\fontsize{10}{11}\selectfont

    \centering
          \caption{Architecture changes for LowFormer edge variant, versions E1/E2/E3 }
    \label{tab:edgevariant_archnumbers}

    \begin{tabular}{c|c|c}
        \Xhline{1.0pt}
         Model  & Changes  & Base model  \\
         \Xhline{0.5pt}

         LowFormer-E1 & a), b), c) & LowFormer-B1.5  \\
         
         LowFormer-E2 & a), b), c) & LowFormer-B3  \\
         LowFormer-E3 &  b) & LowFormer-B3  \\
        
        \Xhline{1.0pt}
     
    \end{tabular}
     \begin{tablenotes}[para,flushleft]
     \footnotesize
     The changes a), b) and c) refer to the enumeration in \Cref{subsec:adapting_lowformer_for_edge}
       \end{tablenotes}

\end{table}


\section{Experiments}
\label{sec:Experiments}

In this section, we present and the discuss the experimental results achieved by the LowFormer family of vision backbones (LowFormer-B0/-/B3), including the edge variants (LowFormer-E1/E2/E3).
For the experiments, we examine model efficiency by measuring GPU throughput, GPU latency and edge device latency. 
Additionally measuring throughput is motivated by its ability to assess how efficiently big quantities of data can be processed. A high throughput is particularly beneficial for tasks such as image retrieval and batched offline processing of video data by detection and segmentation algorithms. It is also crucial to reduce training time \citep{efficientnetv2}.

\bmhead{Protocols for Measuring Execution Time}

In \Cref{tab:measurement_protocols} specifics of the measuring protocols are listed.
We always take the median time per input instance for latency and throughput measurements \citep{fastvit,shvit}.
For latency measurements we always use a batch size of 1, while for GPU throughput we feature a batch size of 200. 
As depicted in \Cref{tab:measurement_protocols}, we measure latency on three edge devices, namely the Nvidia Jetson TX2, the iPhone 13 and the ARM CPU of the Raspberry Pi5 (Arm Cortex A76 processor @ 2.4GHz). 
The amount of iterations differ, because some devices require a higher amount of iterations to retrieve stable results. However we always feature 5 warm-up iterations.
Regarding the iPhone 13 we utilize the CoreML\footnote{\url{https://apple.github.io/coremltools/docs-guides/}} performance tool to retrieve latency results, wherefore we do not have specifics about the amount of iterations.

\begin{table}[bt!]
\fontsize{7}{8}\selectfont
\setlength\tabcolsep{.14cm}
    \centering

        \caption{Specifics of efficiency measurements}
    \label{tab:measurement_protocols}

\begin{tabular}{c|c|c|c}
        \Xhline{1.0pt}
         Device & Metric & Iterations  & Framework  \\
         \Xhline{0.5pt}
        Nvidia A40 GPU & throughput & 100 & PyTorch \\
        Nvidia TITAN RTX GPU & latency & 4000 & TorchScript\tnote{5} \\
        ARM Cortex A76 CPU & latency & 400 & ONNX\tnote{6} \\
    
        iPhone13 & latency & - & CoreML  \\
        Nvidia Jetson TX2 & latency & 200 & ONNX\tnote{6} +TensorRT\tnote{7} \\

        \Xhline{1.0pt}
    \end{tabular}
      \begin{tablenotes}[para,flushleft]
        \footnotesize
        \item[5] \url{https://pytorch.org/docs/stable/generated/torch.jit.optimize_for_inference.html} 
        \item[6] \url{https://github.com/microsoft/onnxruntime}
        \item[7] \url{https://docs.nvidia.com/deeplearning/tensorrt/}
     \end{tablenotes}

\end{table}

\begin{table*}[hbt!]

    \centering
        \caption{Performance on ImageNet-1K validation set }
    \label{tab:imagenetresults}

       \begin{threeparttable}[b]

    \resizebox{\textwidth}{!}{\begin{tabular}{ccccccccc}
        \Xhline{1.5pt}
         \multirow{2}{*}{Model} & \multirow{2}{*}{Venue} & Params  & MACs & GPU Throughput $\uparrow$ & TX2 Latency $\downarrow$ &  ARM CPU $\downarrow$ & Resolution & Top-1  \\
         & & (M) & (M) & (images/s) & (ms) & (ms) & (pixel) & (\%)  \\
        \Xhline{1.5pt}
         
         MobileViG-Ti* \citep{mobilevig} & CVPRW 2023 & 5.3 & 661 & 2500 & \textbf{8.5} & 48.2 & 224 &  75.7  \\
        
         FastViT-T8 \citep{fastvit} & CVPR 2023 & 3.6 & 690 & 1694 & 10.1 & 65.9 & 256 &  75.6  \\

         EfficientMod-xxs \citep{efficientmodulation} & ICLR 2024 & 4.7 & 579 & 2857 & 15.0& 47.5  & 224 &  76.0    \\

         RepViT-M0.9 \citep{repvit} & CVPR 2024 & 5.1 & 816 & 2512 & 11.4 & \underline{40.8} & 224 &  77.4   \\
       
        MobileOne-S2 \citep{mobileone} & CVPR 2023 & 7.8 & 1298 & \underline{2967} & \underline{9.1} & 53.8 & 224 &  77.4  \\

        EdgeViT-XS \citep{edgevit} & ECCV 2022 & 6.8 & 1127 & 2127  & 16.0 & 58.9 & 224 &  77.5  \\
         MobileOne-S3 \citep{mobileone} & CVPR 2023 & 10.1 & 1895 & 2433 & 11.8 & 74.2 & 224 &  78.1 \\
         MobileViG-S* \citep{mobilevig} &  CVPRW 2023 & 7.3 & 983 & 1724 & 12.3 & 73.9 & 224 &  78.2  \\
        
        EfficientMod-xs \citep{efficientmodulation} & ICLR 2024 & 6.6 & 773 & 2352 & 17.7 & 53.8 & 224 &  \underline{78.3}   \\

           \rowcolor{lightergray} LowFormer-B0 (ours)& &  14.1 & 944 & \textbf{5988} & \textbf{8.5} & \textbf{39.1} & 224 &  \textbf{78.4 }  \\

         \Xhline{1.5pt}

      FastViT-T12 \citep{fastvit} & CVPR 2023 & 6.8 & 1400 & \underline{2054} & 14.5 & 110.4 & 256 &  79.1  \\

           RepViT-M1.1 \citep{repvit} & CVPR 2024 & 8.2 & 1338 & 1941 & \underline{13.5} & \underline{63.5} & 224 &  \underline{79.4}    \\
        
        MobileOne-S4 \citep{mobileone} & CVPR 2023 & 14.8 & 2978 & 1550 & 18.6 & 122.9 & 224 &  \underline{79.4}  \\
        
         \rowcolor{lightergray} LowFormer-B1 (ours) & & 17.9 & 1410 & \textbf{4237} & \textbf{11.7} & \textbf{59.1} & 224 &  \textbf{79.9 }  \\

 \Xhline{1.5pt} 
 
      EfficientFormerV2-S2 \citep{efficientformerv2} & ICCV 2023 & 12.6 & 1250 & 468  & 19.9 & 102.3 & 224 &  80.4  \\

       FastViT-SA12 \citep{fastvit} & CVPR 2023 & 10.9 & 1943 & 1075 & \textbf{17.5} & 136.4 & 256 &  80.6  \\

      EdgeViT-S \citep{edgevit} & ECCV 2022 & 11.1 & 1910 & 1449  & 24.6 & \textbf{99.2} & 224 &  81.0  \\

        EfficientMod-s \citep{efficientmodulation} & ICLR 2024 & 12.9 & 1402 & \underline{1381} & 30.5 & 105.6 & 224 &  \underline{81.0}   \\
    
        RepViT-M1.5 \citep{repvit} & CVPR 2024 & 14.0 & 2276 & 1146 & 23.0 & 113.2 & 224 & \textbf{ 81.2}  \\

        \rowcolor{lightergray} LowFormer-B1.5 (ours) & & 33.9 & 2573 & \textbf{2739} &\underline{18.1} & \underline{111.6} & 224 &  \textbf{81.2}  \\
         
        \Xhline{1.5pt}

        FFNet-1 \cite{ffnet} & arXiv 2024 & 13.8 & 3000 & 1090 & 30.4 & 242.1 & 256 &  81.3 \\

        BiFormer-T \cite{biformer} & CVPR2023 & 13.1 & 2200 & 729 & 61.3 & 523.9 & 224 &  \underline{81.4}  \\

         \rowcolor{lightergray} LowFormer-B2 (ours) & & 45.0 & 3689 & \textbf{2227} & \textbf{21.6} & \textbf{144.2} & 224 &  \textbf{81.6}  \\

        \Xhline{1.5pt}

        SMT-T \citep{smtbackbone} & ICCV2023 & 11.5 & 2400 & \underline{770} & 50.3 & \textbf{195.6} & 224 &  82.2  \\
        
        RepViT-M2.3 \citep{repvit}& CVPR 2024 & 22.9 & 4520 & 642 & 40.6 & 227.0 & 224 &  82.5    \\
       FastViT-SA24 \citep{fastvit}& CVPR 2023 & 20.6 & 3769 & 606 & \underline{30.9} &   273.7 & 256 & \underline{82.6}   \\

        \rowcolor{lightergray}  LowFormer-B3$_{r192}$ (ours)& & 57.1 & 4479 & \textbf{1562} & \textbf{30.0} & \underline{198.6} & 192 &  \textbf{82.7} \\
         
         \Xhline{1.5pt}

       iFormer-S \citep{inceptiontransformer}& NeurIPS 2022 & 19.9 & 4825 & 555  & 51.2 & \underline{270.6} & 224 &  83.4  \\

        FastViT-SA36 \citep{fastvit} & CVPR 2023 & 30.4 & 5595 & 429 &  \underline{44.4} & 399.8 & 256 & 83.6    \\
       SMT-S \citep{smtbackbone} & ICCV2023 & 20.5 & 4700 & 418 & 96.4 & 397.8 & 224 &  \underline{83.7 } \\
        
        BiFormer-S \citep{biformer} & CVPR2023 & 26.0 & 4500 & 348 & 130.2 & 1134.1 & 224 &  \textbf{83.8 } \\

       \rowcolor{lightergray} LowFormer-B3 (ours)& & 57.1 & 6098 & \textbf{1162} & \textbf{32.5} & \textbf{238.4} & 224 &  83.6  \\

        \Xhline{1.5pt}

    \end{tabular}}
   
    \begin{tablenotes}[para,flushleft]
        \footnotesize
        The table is divided into different groups, determined by similar 
        top-1 accuracy (bold horizontal   lines  separate groups).
    Values in bold are the best results for each \\  group  and column,  while underlined results refer to  the second best. For models marked  with  *, only  distilled model results are publicly available. LowFormer \\ models (highlighted in gray) achieve superior speed accuracy trade-offs in terms of GPU throughput, TX2 latency and ARM CPU latency
     \end{tablenotes}
  
  \end{threeparttable}

\end{table*}

\subsection{ImageNet-1K Classification}
\label{subsec:imagenetclass}

\bmhead{Settings}
We conduct image classification experiments on ImageNet-1K \citep{imagenet}, which includes 1.28M training and 50K validation images for 1000 categories. 
All models were trained from scratch using a similar setting as \citep{efficientvit} and featuring an input resolution of 224. We also trained for a total of 320 epochs using AdamW \citep{adamwpaper} optimizer and a learning rate of $10^{-3}$, however we use a batch size of 512. 
As learning rate scheduler we use cosine decay \citep{cosinelrpaper} and 20 warm-up epochs with a linear schedule. We also feature the multi-scale learning from \citep{efficientvit}.
We trained LowFormer-B3 with a batch size of 2400 and a base learning rate of $3\times10^{-3}$. For LowFormer-B2 we had a batch size of 850 and a base learning rate of $8.3\times10^{-4}$.

\begin{figure}
 
   \includegraphics[width=\linewidth]{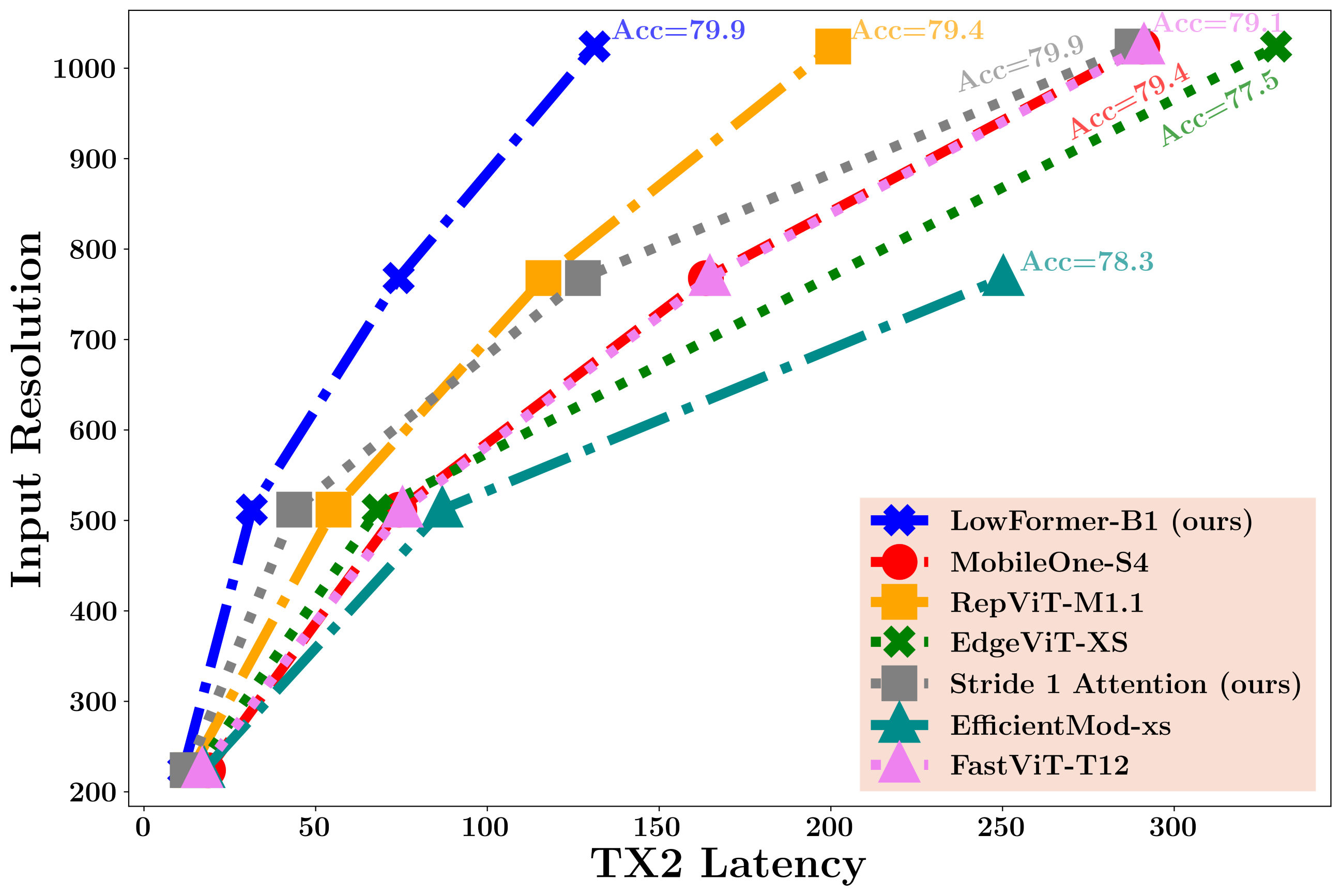}
  \caption{Impact of input resolution on Jetson TX2 latency for LowFormer-B1 (ours), LowFormer-B1 without downsampling in the \lowformeratt { } layers (see \Cref{subsec:ablation}), MobileOne-S4 \citep{mobileone}, RepViT-M1.1 \citep{repvit}, EdgeViT-XS \citep{edgevit}, EfficientMod-xs \citep{efficientmodulation} and FastViT-T12 \citep{fastvit}. "Acc" refers to ImageNet top-1 accuracy. LowFormer-B1 demonstrates remarkable efficiency when operating at a higher input resolution, outperforming compared state-of-the-art architectures}
  \label{fig:resolution_scaling_tx2}
\end{figure}

\bmhead{Results}
In \Cref{tab:imagenetresults}, we evaluate the speed accuracy trade-off of the compared models. For this we measure ImageNet \citep{imagenet} top-1 accuracy and assess efficiency based on  GPU throughput, Jetson TX2 latency and ARM CPU latency.
The base models of LowFormer (LowFormer-B0 to B3) achieve a superior speed accuracy trade-off, outperforming most compared models in all three efficiency metrics.
LowFormer-B0, has a slightly higher top-1 accuracy than EfficientMod-xs \citep{efficientmodulation}, twice the throughput, half of its TX2 latency and executes 38 \% faster on ARM CPU.
Although BiFormer-T \citep{biformer} has 40\% less MACs than LowFormer-B2$_{r224}$, our LowFormer has a 305\% higher throughput and only a third of its TX2 and ARM CPU latency. At the same time it slightly outperforms BiFormer-T by 0.2\% top-1 accuracy.
The largest base model of LowFormer, LowFormer-B3, surpasses FastViT-SA36 \citep{fastvit} in efficiency, achieving nearly three times the GPU throughput and running 36\% faster on the TX2, and 67\% faster on an ARM CPU. Both models share the same top-1 accuracy.
The hardware efficient design of LowFormer base models not only execute MACs more efficiently but also leads to higher top-1 accuracies, while achieving improved efficiency.

\begin{table*}[hbt!]
    \centering
        \caption{Ablation study of LowFormer-B1, featuring singular changes to the original model }
    \label{tab:ablation}

       \begin{threeparttable}[b]

    \resizebox{\textwidth}{!}{\begin{tabular}{c|cc|ccc|c}
        \Xhline{1.5pt}
         \multirow{2}{*}{Model version}  & Params &  MACs & GPU Throughput $\uparrow$ & TX2 Latency $\downarrow$ & ARM CPU Latency $\downarrow$ & Top-1   \\
         & (M) & (M) & (images/s) & (ms) & (ms) & (\%)  \\
         \Xhline{1.0pt}

            unfused MBConv & 12.4 & 716 & 3558 {\scriptsize  \textbf{(-16\%)}} & 12.8 {\scriptsize \textbf{(+9\%)}} & \textbf{55.5} {\scriptsize \textbf{(-6\%)}}  & 79.1 {\scriptsize \textbf{(-0.8)}}  \\
        
        relu-linear att  & 14.15 & 1210 & 3367 {\scriptsize  \textbf{(-20\%)}} & 12.7 {\scriptsize \textbf{(+8\%)}} & 64.8  {\scriptsize \textbf{(+9\%)}} & 79.6 {\scriptsize \textbf{(-0.3)}} \\
        
        original MHSA  & 16.8  & 1460 & 3590 \textbf{(-15\%)} & 12.5 \textbf{(+7\%)} & 61.2  {\scriptsize \textbf{(+4\%)}} & 79.8 \textbf{(-0.1)} \\
        
        high-res attention & 17.65 & 1494 & 3759 {\scriptsize  \textbf{(-11\%)}} &  13.4 {\scriptsize \textbf{(+14\%)}} &  64.0  {\scriptsize \textbf{(+8\%)}} & \textbf{79.9} {\scriptsize \textbf{(+0.0)}}\\

        no channel compr. & 20.68 & 1650 & 3921 {\scriptsize  \textbf{(-7\%)}} &  13.4 {\scriptsize \textbf{(+14\%)}}& 63.1  {\scriptsize \textbf{(+7\%)}} & \textbf{79.9} {\scriptsize \textbf{(+0.0)}}\\

        \Xhline{1.0pt}
        
        Baseline (B1) & 17.94 & 1410 &  \textbf{4237} & \textbf{ 11.7} & 59.1 & \textbf{79.9}  \\
        
        \Xhline{1.5pt}
     
    \end{tabular}}
    
     \begin{tablenotes}[para,flushleft]
        \footnotesize
        Bold entries mark the best in its column. The LowFormer architecture design elements consistently demonstrate superior efficiency while achieving higher \\ ImageNet  accuracy
     \end{tablenotes}
      
      \end{threeparttable}

\end{table*}

\bmhead{Resolution Scaling}
A model's efficiency at increased input resolution is critical, as many downstream tasks run the backbone on high-resolution inputs \citep{Detr,segmentanything}. As shown in \Cref{tab:highreshighchan}, model efficiency can vary in response to higher input resolutions, depending on the architecture.
Therefore, we examine in  \Cref{fig:resolution_scaling_tx2} how increasing input resolution affects latency on the Jetson TX2 for LowFormer-B1 and other approaches.
LowFormer-B1 outperforms depicted models in top-1 accuracy on ImageNet-1K and at the same time remains considerably faster, independent of the input resolution. 
The model ``Stride 1 Attention'' refers to an ablation of LowFormer-B1, discussed in \Cref{subsec:ablation}.

\subsection{Ablation Study of LowFormer Base Models}
\label{subsec:ablation}
In \Cref{tab:ablation} we ablate our model design decisions. 
We revert a singular design decision of LowFormer-B1 to demonstrate the impact of that change on accuracy, GPU throughput and TX2 latency. 
The featured ablations are the following:

\begin{itemize}
    \item We replace all fused MBConv blocks with the unfused version.

     \item We replace our \lowformeratt { } with ReLU linear attention from \citep{efficientvit} in order to compare our attention approach with other recent adaptations.

    \item We omit the downscaling of the feature maps for the \lowformeratt{ }(high-res attention).
    \item We remove the channel compression done during the projection in \lowformeratt.
    \item We replace \lowformeratt { }with the original MHSA proposed by  \citep{attentionisallyouneed}.
    
\end{itemize}

\subsubsection{Ablation Results}
\bmhead{Unfused MBConv}
Replacing the fused MBConv with the unfused one results in a 16\% lower GPU throughput and 10\% higher TX2 latency (see \Cref{tab:ablation}). On the other side ARM CPU latency improves by 6\%, however top-1 accuracy drops significantly by 0.8\%  
As we can see, next to a mostly improved execution time, fusing the MBConv can increase performance significantly. \\

\bmhead{ReLU Linear Attention}
Applying ReLU linear attention from \citep{efficientvit} on the other side results in 0.3\% reduced top-1 accuracy, as well as \~8\% higher latency and 20\% lower GPU throughput, showing the benefit of \lowformeratt.
\bmhead{Downsampling in \lowformeratt}
When we remove the downsampling for the \lowformeratt, top-1 accuracy stays the same, but GPU throughput and latency worsen significantly. In \Cref{fig:resolution_scaling_tx2} we can see that for higher input resolutions the latency difference multiplies. For input resolution 1024$\times$1024 for example, latency is increased by 220\% on the Jetson TX2.
\bmhead{Channel Compression}
Removing the channel compression of \lowformeratt { } during the projection phase does not improve accuracy, even though the SDA operates on a lower channel dimension with  compression. However throughput and latency worsen significantly without channel compression.  

\bmhead{MHSA}
When removing channel compression and convolutions from \lowformeratt { }(no downsampling), we revert to the original MHSA. As shown in \Cref{tab:ablation}, this not only reduces top-1 accuracy,
but also increases model latency significantly. This difference multiplies, when input resolution is increased as shown in \Cref{tab:attentionspeedexp}.

\bmhead{Summary}
The reduced dimensions on which we apply the attention operation (channel compression and resolution reduction) have no effect on top-1 accuracy, but improve the efficiency of the model significantly. The additional convolutions improve accuracy (original MHSA fairs worse in \Cref{tab:ablation}), but only mildly worsen execution time (see \Cref{tab:attentionspeedexp}).
In conclusion \lowformeratt { }is both more efficient and effective than the original MHSA \citep{attentionisallyouneed}, with significantly better scalability for increased input resolution.

\begin{table*}[hbt!]
    \centering
        \caption{Efficiency comparison of LowFormer on several computing devices with modified versions of the original architecture, where the multi-layer-perceptron (MLP) or the \lowformeratt { }(Att) is removed, or the depth of the last two stages is reduced }
    \label{tab:model_choices}

       \begin{threeparttable}[b]
    
    \resizebox{\textwidth}{!}{\begin{tabular}{ccccccccc}
        \Xhline{1.5pt}
         \multirow{2}{*}{Model} & MLP  & Original & Att & GPU Throughput & GPU Latency & TX2 Latency & iPhone 13 &   Top-1  \\
         & & Depth &  & (images/s)  & (ms) & (ms) & (ms)  & (\%)  \\
        \Xhline{1.5pt}
    
        \rowcolor{lightergray} LowFormer-B0 & \cmark & \cmark & \cmark & 5988 & 2.9 & 8.5 & 1.5    &  78.4  \\
        \rowcolor{lightergray} LowFormer-B1 &  \cmark &\cmark & \cmark & 4237 & 4.0 & 11.7 & 1.8    &  79.9  \\
        
        B1\_mlpless & \xmark & \cmark & \cmark  & 6067  & 2.8 & 8.5 & 1.5    &  78.8  \\
        B1\_mlpless\_shallow & \xmark & \xmark & \cmark  & 8254 & 1.8 & 5.9 & 1.3    &  77.2  \\
         \Xhline{1.5pt}         
        \rowcolor{lightergray} LowFormer-B1.5 &  \cmark & \cmark & \cmark & 2739  & 4.8 &  2.8   & 111.6 &  81.2  \\
        B1.5\_mlpless & \xmark & \cmark & \cmark & 4019  & 3.3 & 13.2 & 2.4   &  80.7  \\
        B1.5\_mlpless\_shallow & \xmark & \xmark & \cmark & 5268  & 2.4 & 9.7 & 2.0   &  79.7  \\
        B1.5\_conv\_shallow (E1) & \xmark & \xmark  & \xmark & 6337  & 1.0 & 6.2 & 1.6    &  78.8  \\

         \Xhline{1.5pt} 
        \rowcolor{lightergray} LowFormer-B3 & \cmark & \cmark & \cmark & 1162  & 5.2 & 32.5 & 4.5   &  83.6  \\
        B3\_mlpless (E3) & \xmark & \cmark & \cmark & 1566  & 3.6 & 25.0 & 3.6    &  83.0  \\
        B3\_mlpless\_shallow & \xmark & \xmark & \cmark & 1848  & 2.7 & 19.6 & 2.8    &  82.2  \\
       B3\_conv\_shallow (E2) &  \xmark & \xmark & \xmark & 2070  & 1.5 & 14.7 & 2.5    &  81.6  \\

        \Xhline{1.5pt}

    \end{tabular}}
    
\begin{tablenotes}[para,flushleft]
        \footnotesize
The highest efficiency increase can be achieved by removing the attention operation. The connotations E1,E2,E3 in braces  refer  to the proposed edge GPU variants. \\ LowFormer base models are highlighted in gray
         \end{tablenotes}
  
  \end{threeparttable}

\end{table*}

\subsection{Comparison of LowFormer Edge GPU Variants with the Base Models}
\label{sec:edgebone}

In \Cref{subsec:adapting_lowformer_for_edge} we presented the three derivations LowFormer-E1/E2/E3 from the original LowFormer base models (B1.5 and B3). 
In the following part we will first experimentally justify why these specific changes were chosen by analyzing the effect of each change on efficiency and accuracy. Then we will put the LowFormer edge variants into perspective with the best competing models of \Cref{tab:imagenetresults}, based on their speed accuracy trade-off on the Jetson TX2.

\subsubsection{Efficiency Analysis of Attention, MLP and Depth}
\label{subsec:eff_analysis_edgebones}
In \Cref{subsec:adapting_lowformer_for_edge} we presented three possible changes to the LowFormer architecture, namely removing the MLP, reducing the depth and removing the \lowformeratt. In \Cref{tab:model_choices} we show the effect of each of these changes cumulatively. First we remove the MLP ("mlpless"), then we additionally reduce the depth ("shallow") and at last we also remove the \lowformeratt { }("conv"), leaving only convolutions in the architecture. 
We use LowFormer B1, B1.5 and B3 as baselines for the modifications. All derivations have the same hyperparameter setting during training as their corresponding base version.

\bmhead{Removing Attention}
From the model B1.5\_conv\_shallow we can see the enourmous penalty attention can have on latency. It has 42\% of the GPU latency and 64\% of the TX2 latency of B1.5\_mlpless\_shallow, but loses less than 1\% top-1 accuracy. Both models only differ in that B1.5\_conv\_shallow does not feature the \lowformeratt.
The difference becomes more apparent if you compare B1.5\_conv\_shallow to b1\_mlpless, which has the same top-1 accuracy, but fares far worst regarding GPU and TX2 latency.
\bmhead{Removing MLP}
On the other side, omitting the MLP can also be beneficial, although not as pivotally. 
B1\_mlpless has a 0.4\% higher top-1 accuracy than LowFormer-B0, while being slightly faster in terms of latency and throughput.
\bmhead{Reducing Model Depth}
Reduction of the model depth especially improves latency, while its impact on throughput is less pronounced. For example B1.5\_mlpless\_shallow achieves a 0.9\% higher top-1 accuracy compared to B1\_mlpless, while having an improved GPU latency and a slightly worse throughput.

\bmhead{Summary}
Removing \lowformeratt, the MLP and reducing the model depth can significantly improve the speed accuracy trade-off, especially in terms of latency on GPU and edge GPU (Jetson TX2).
These three modifications are combined in LowFormer-E1 and LowFormer-E2, yielding a substantial efficiency gain  with only a minimal  accuracy drop.
In LowFormer-E3 only the MLP is removed, yet it still achieves a notable efficiency boost, while maintaining higher top-1 accuracy without requiring additional model scaling methods like width scaling \citep{efficientnet}, which become increasingly inefficient with higher model capacity.

\begin{table*}[hbt!]
\fontsize{10}{11}\selectfont
    \centering
    \caption{Efficiency comparison between the LowFormer edge GPU variants and the best competing models of \Cref{tab:imagenetresults} }
    \label{tab:final_comparison}

       \begin{threeparttable}[b]
   
    \resizebox{\textwidth}{!}{\begin{tabular}{ccccccccc}
        \Xhline{1.5pt}
         \multirow{2}{*}{Model} & MACs & GPU Throughput & GPU latency & TX2 Latency & iPhone 13   & Top-1  \\
         & (M) &   (images/s) & (ms) & (ms) & (ms) & (\%)  \\
        \Xhline{1.5pt}
    
    EfficientMod-xxs \citep{efficientmodulation}& 579 & 2857 & 2.1 & 15.0 & 1.71   & 76.0 \\
    
    EdgeViT-XS \citep{edgevit}& 1127 & 2127 & 2.7 & 16.0 & \underline{1.5}  & 77.5 \\
    
    MobileOne-S3 \citep{mobileone} & 1895 & 2433  & \underline{1.0} & 11.8 & \textbf{1.2} & 78.1 \\
    EfficientMod-xs \citep{efficientmodulation}& 773 & 2352 & 2.5 & 17.7 & 2.2   & 78.3 \\
    
         LowFormer-B0 (ours) & 944 & \underline{5988} & 2.9 & \underline{8.5} & \underline{1.5}  & \underline{78.4} \\
       \rowcolor{lightergray} LowFormer-E1 (ours) & 1350 & \textbf{6337} & \textbf{1.0} & \textbf{6.2} & 1.7   & \textbf{78.8 }\\

\Xhline{1.5pt}    
       
         FastViT-SA12 \citep{fastvit} & 1943 & 1075 & \underline{1.7} & \underline{17.5} & \underline{1.6}   & 80.6 \\
         EfficientMod-s \citep{efficientmodulation}& 1402 & 1381 & 3.8 & 30.5 & 2.6   & 81.0 \\
       RepViT-M1.5 \citep{repvit} & 2276 & 1146 & 4.2 & 23.0 & \textbf{ 1.5}   & 81.2 \\

        LowFormer-B1.5 (ours) & 2573  & \textbf{2739}  & 4.8 & 18.1 & 2.8  & \underline{81.2} \\
       LowFormer-B2 (ours) & 3689 & \underline{2227} & 4.8 & 21.6 & 3.5   &\textbf{ 81.6} \\
       
      \rowcolor{lightergray} LowFormer-E2 (ours)  & 3800 & 2070 & \textbf{1.5} & \textbf{14.7} & 2.5   & \textbf{81.6} \\
       
\Xhline{1.5pt}        
       FastViT-SA24 \citep{fastvit}& 3769 & 606 & \textbf{3.0} & 30.9 & \underline{2.6}   & 82.6 \\
        RepViT-M2.3 \citep{repvit} & 4520 & 642 & 5.5 & 40.6 & \textbf{2.4}   & 82.5 \\

        LowFormer-B3$_{r192}$ (ours) & 4479 & \underline{1562} &  5.5 & \underline{30.0} & 4.5   & \underline{82.7} \\
     \rowcolor{lightergray}  LowFormer-E3 (ours)  & 5350 & \textbf{1566} & \underline{3.6} & \textbf{25.0} & 3.6   & \textbf{83.0} \\
        \Xhline{1.5pt}

    \end{tabular}}
    
    \begin{tablenotes}[para,flushleft]
        \footnotesize
        LowFormer-E1/E2/E3 (highlighted in gray) consistently rank among the most efficient models in terms of GPU throughput, GPU latency, and TX2 latency
\end{tablenotes}
  
  \end{threeparttable}

\end{table*}

\subsubsection{Evaluation of Edge Optimization}
To put our LowFormer edge GPU variants in perspective, we compare them in \Cref{tab:final_comparison} to the highest competing models from \Cref{tab:imagenetresults} based on their respective Jetson TX2 latency.   
 \bmhead{GPU Throughput \& TX2 Latency }
 LowFormer-E1/E2/E3 consistently achieve a better throughput and TX2 latency than all compared state-of-the-art models with a similar or lower top-1 accuracy, including the LowFormer base models (B0/B1.5/B2/B3). LowFormer-E1 for example has half of the TX2 latency of MobileOne-S3 \citep{mobileone}, while having 0.7\% higher accuracy. LowFormer-E2 has a similar latency compared to FastViT-T12 \citep{fastvit}, but scores 2.5\% higher in ImageNet top-1 accuracy.

 \bmhead{GPU Latency}
 LowFormer-E1/E2 similarly outperform all compared models in GPU latency, however LowFormer-E3 fares considerably worse in that regard. This is mainly to the fact, that it still makes use of the attention operation, which has a negative impact on GPU latency (as mentioned in \Cref{subsec:eff_analysis_edgebones}). Nevertheless it is considerably more efficient when compared to LowFormer-B3$_{r192}$.
\bmhead{Mobile Latency}
Regarding mobile execution on the iPhone 13, the edge variant do not give a consistent speed-up, when compared with the LowFormer base versions. Moreover, architectures like RepViT and FastViT \citep{repvit, fastvit} achieve superior performance on iPhone 13. This is in part because the edge variants and base models feature a higher amount of MACs, which
the mobile compute hardware cannot effectively parallelize.

 \bmhead{Summary}
 LowFormer-E1, E2 and E3 lead the table (\Cref{tab:final_comparison}) in terms of GPU throughput, GPU latency and Jetson TX2 latency. However, the optimizations from \Cref{subsec:adapting_lowformer_for_edge} do not translate well to mobile execution on the iPhone 13 NPU and GPU. Despite this, the edge GPU variants achieve a speed-up of up to 3×, compared to the LowFormer base models.

\subsection{Application to Downstream Tasks}
For a fair comparison, we compare models of similar size with each other and select those that achieve the best performance in the respective benchmark.

\begin{table*}[t]
        \fontsize{8}{9}\selectfont
\setlength\tabcolsep{.14cm}
    \centering

     \begin{adjustbox}{max width=1.0\textwidth, center}
  
       \begin{threeparttable}[b]
            \caption{Evaluation on transfer learning classification datasets }
        \label{tab:imagenetdownstream}

    \begin{tabular}{c|c|ccc}
        \Xhline{1.5pt}
         \multirow{2}{*}{Model}  & GPU Throughput $\uparrow$ & Flowers & Cars & Pets \\
          & (images/sec) & Top-1 (\%) & Top-1 (\%) & Top-1 (\%) \\
          \Xhline{1.0pt}

            \Xhline{1.0pt}

        ViT-L/16 \citep{ViT} & 36 & 89.7 & - & 93.6 \\
        ViT-B/16 \citep{ViT}  & 117 &  89.5 & -  & 93.8 \\
       
        TNT-S \citep{han2021transformertnt} & 141 & 98.8  & -  & 94.7 \\       
        DeiT-B \citep{deitdistill} & 114 & \textbf{98.9} & 93.9 & - \\
        EfficientNetV2-M \citep{efficientnetv2} & 277 & 98.5 & \textbf{94.6} & - \\
       
        CeiT-S \citep{ceitbackbone} & 260 & 98.6  & 94.1 & 94.9 \\

        \rowcolor{lightergray} LowFormer-B3 (ours)  & \textbf{424} & \textbf{98.9} & 94.4 & \textbf{95.0} \\

     \Xhline{1.5pt}
    \end{tabular}
     \begin{tablenotes}[para,flushleft]
        \footnotesize
        All models are finetuned and evaluated on resolution 384×384. Best Results for each column are marked bold. LowFormer model is highlighted in gray
  \end{tablenotes}
  
  \end{threeparttable}
 \end{adjustbox}

\end{table*}

\subsubsection{Image Classification}
We assess LowFormer's transfer learning capabilities by evaluating its  performance when finetuned on smaller image classification datasets.
We feature three datasets, namely Oxford-IIIT-Pets \citep{petsdataset}, Stanford Cars \citep{stanfordcars} and Oxford-102 Flowers \citep{oxfordflowers}, following \citep{efficientnetv2,ViT,han2021transformertnt}.

\bmhead{Settings}
For finetuning LowFormer on the classification datasets, we maintained a setup similar to that used for ImageNet training (see \Cref{subsec:imagenetclass}). 
However we increased training and evaluation resolution to 384×384, applied a batch size of 512, a base learning rate of $2.5\times10^4 $ and removed weight decay, following previous procedures for transfer learning datasets \citep{efficientnetv2,ViT}. 

We train for 360 steps on the train splits of Oxford-IIIT-Pets \citep{petsdataset}, 800 steps on Oxford-102 Flowers \citep{oxfordflowers}, and 3200 steps on Stanford Cars \citep{stanfordcars}.

\bmhead{Results}
In \Cref{tab:imagenetdownstream}, we compare the evaluation results of LowFormer-B3 against both convolutional and transformer-based approaches. GPU throughput  is measured at a resolution of 384×384.
The LowFormer models mostly achieve superior results across all three datasets \citep{petsdataset,oxfordflowers,stanfordcars}, while maintaining equal or lower GPU throughput. 
The ViT \citep{ViT} models, for instance, fall significantly behind in both efficiency and accuracy, whereas CeiT-S \citep{ceitbackbone} achieves accuracy results  closer to LowFormer-B3 but with only half the GPU throughput.

\subsubsection{Object Detection }
\label{subsubsec:od_lowformer}
We show the applicability of the LowFormer architecture for object detection.  
Backbone GPU throughput and Jetson TX2 latency measurement in \Cref{tab:objectdetection} are conducted using an input resolution of 512$\times$512 \citep{shvit, fastvit, fat}.

\bmhead{Settings}
We plug the pretrained LowFormer base models (B0/B1/B2/B3) into the RetinaNet framework \citep{retinanet} and utilize COCO 2017 \citep{cocopaper} for training and evaluation.
We train the LowFormer base models for 12 epochs (1x schedule) and following \citep{efficientvit,fat} regarding all hyperparameters.
As evaluation metric we use mean average precision (mAP)\footnote[8]{Mean average precision (mAP) is commonly abbreviated as AP in many publications that evaluate on COCO. Within the context of COCO evaluation, AP always refers to mAP.} \citep{edgevit,fat}.

\begin{table*}[t]
    \centering
        \caption{Comparison results on object detection on COCO 2017 \citep{cocopaper} using RetinaNet \citep{retinanet} head}
    \label{tab:objectdetection}

       \begin{threeparttable}[b]

    \resizebox{\textwidth}{!}{\begin{tabular}{c|cc|cccccc}
        \Xhline{1.5pt}
         \multirow{2}{*}{Backbone}  & GPU Throughput & TX2 Latency & mAP &  mAP$_{50}$ & mAP$_{75}$ & mAP$_{s}$ & mAP$_{m}$ & mAP$_{l}$ \\
         & (images/s) & (ms) & (\%) & (\%) & (\%) & (\%) & (\%) &  (\%)  \\
         \Xhline{1.0pt}

        MobileNetV3 \citep{mobilenetv3}  & 862 & 19.7 & 29.9 & 49.3 & 30.8 & 14.9 & 33.3  & 41.1 \\

        MobileNetV4-Conv-M \citep{mobilenetv4}  & 517 & 27.4 & 32.6 & - & - & - & -  & - \\

        EfficientViT-M4 \citep{efficientvitmemory}  & \textbf{1700}  & \textbf{17.6} & 32.7 & 52.2 & 34.1 & 17.6 & 35.3 & 46.0  \\ 
        PVTv2-B0  \citep{pvtv2} & 355 & 96.0 & 37.2 & 57.2 & 39.5 & \textbf{23.1} & 40.4 & 49.7 \\

        \rowcolor{lightergray}  LowFormer-B0 (ours)  & 1190  & 22.4 & \textbf{38.6 }& \textbf{59.1} & \textbf{40.9} & 21.8 & \textbf{41.8 }& \textbf{51.7}  \\
        \Xhline{1.0pt}
        
        EdgeViT-XXS \citep{edgevit}  & 518 & 51.8 & 38.7 & 59.0 & 41.0 & \textbf{22.4} & 42.0  & 51.6   \\
        
         \rowcolor{lightergray} LowFormer-B1 (ours)  & \textbf{840} & \textbf{31.6} & \textbf{39.4 }& \textbf{59.8} & \textbf{41.7 }& \textbf{22.4} & \textbf{42.9} & \textbf{52.4}  \\
        \Xhline{1.0pt}
     
        FAT-B0 \citep{fat}  & 232 &  94.2 & 40.4 & 61.6 & 42.7 & 24.0 & 44.3 & 53.1  \\
        
        EdgeViT-XS \citep{edgevit}  & 400 & 68.1 & 40.6 & 61.3 & 43.3 & 25.2 & 43.9  & 54.6   \\

        PVTv2-B1  \citep{pvtv2} & 215 & 268.8 & 41.2 & 61.9 & 43.9 & \textbf{25.4} & 44.5 & 54.3  \\
       
        \rowcolor{lightergray} LowFormer-B2 (ours) & \textbf{450} & \textbf{63.3} & \textbf{41.4} & \textbf{62.2} & \textbf{ 44.1 } &  24.5 &  \textbf{45.1 }& \textbf{55.5}  \\

        \Xhline{1.0pt}

        FAT-B1 \citep{fat}  & 174 & 125.0  & 42.5 & 64.0 & 45.1 & 26.9 & 46.0 & \textbf{56.7}  \\
        
         \rowcolor{lightergray} LowFormer-B3 (ours) & \textbf{245} & \textbf{109.0} & \textbf{43.1} & \textbf{ 64.5} &  \textbf{45.9} &  \textbf{27.1} &  \textbf{47.1} & \textbf{ 56.7}  \\
     \Xhline{1.5pt}
    \end{tabular}}
    
     \begin{tablenotes}[para,flushleft]
        \footnotesize
         LowFormer base models (B0, B1, B2, B3) are able to outperform all compared models in speed accuracy trade-off. Backbone GPU  throughput and TX2 latency \\ are measured under resolution of 512$\times$512. LowFormer base models are highlighted in gray
     \end{tablenotes}
  
  \end{threeparttable}

\end{table*}

\begin{table*}[t]

        \fontsize{7}{8}\selectfont
\setlength\tabcolsep{.14cm}
    
        \begin{flushleft}
    \begin{minipage}[t]{0.1\textwidth}
    \raggedleft

       \begin{threeparttable}[t]
       \caption{Results on semantic segmentation, using Semantic FPN \citep{semanticfpn} }
     \label{tab:segmentation}
     
    \begin{tabular}{c|cc|c}
        \Xhline{1.5pt}
         \multirow{2}{*}{Backbone}  & GPU Throug. & TX2 Lat.  & mIoU    \\
         & (images/s) &  (ms) & (\%) \\
         \Xhline{1.0pt}

        ResNet50  \citep{resnetpaper} & 271 & \underline{45.9} & 36.7 \\
        PVTv2-B0  \citep{pvtv2} & 355 & 96.0 & 37.2 \\
        FastViT-SA12  \citep{fastvit} & 265 & 62.0 & \underline{38.0} \\
        EdgeViT-XXS \citep{edgevit}  & \underline{518} & 51.8 & \textbf{39.7}  \\
        \rowcolor{lightergray} LowFormer-B1 (ours) & \textbf{840} &  \textbf{31.6} & \textbf{ 39.7} \\
        \Xhline{1.0pt}

        RepViT-M1.1  \citep{repvit} & \underline{404} & \textbf{55.0} & 40.6 \\
        FastViT-SA24  \citep{fastvit} & 151 & 109.9 & 41.0 \\
        EdgeViT-XS \citep{edgevit}  & 400  & 68.1 & 41.4  \\
        FAT-B0 \citep{fat}  & 232 & 94.2 & 41.5 \\
                
        EfficientFormerV2-S2 \citep{efficientformerv2}  & 182 & 85.3 & 42.4 \\
        
        PVTv2-B1  \citep{pvtv2} & 215 & 268.8 & \underline{42.5} \\

        \rowcolor{lightergray} LowFormer-B2 (ours) & \textbf{450} & \underline{63.3} & \textbf{ 42.8} \\
        \Xhline{1.0pt}
        
        FAT-B1 \citep{fat}  & 174 & \underline{125.0}  & 42.9 \\
        RepViT-M1.5  \citep{repvit} & \underline{238} & 217.4 & \underline{43.6} \\
        
        FastViT-MA36  \citep{fastvit} & 86 & 208.4 & \textbf{44.6} \\
       \rowcolor{lightergray} LowFormer-B3 (ours) & \textbf{245}  & \textbf{109.0} & \textbf{ 44.6} \\
    
        \Xhline{1.5pt}
     
    \end{tabular}
    \begin{tablenotes}[para,flushleft]
        \footnotesize
        LowFormer models achieve superior speed mIoU trade-offs. Backbone GPU throughput and TX2 latency are measured under resolution of 512$\times$512. Results are grouped by mIoU. Bold marks the best results in each group and column, underline refers to the second best. LowFormer models are highlighted in gray
     \end{tablenotes}
  
  \end{threeparttable}

\end{minipage} \hspace{7.0cm} \begin{minipage}[t]{0.1\textwidth} 

 \begin{threeparttable}[t]
      \caption{Image retrieval results on GPR1200 \citep{gpr1200} benchmark }
    \label{tab:gpr1200results}
\begin{tabular}{c|cc|c}
        \Xhline{1.5pt}
         \multirow{2}{*}{Backbone}  & Resol. &GPU Throug.   & mAP    \\
         & (pixel) &(images/s) &   (\%) \\
         \Xhline{1.0pt}

        EfficientViT-M5 \citep{efficientvitmemory} &  224 &  5681  & 31.9  \\
        SHViT-S4 \citep{shvit} &  256 &  4255  &  35.7  \\
        FastViT-T8 \citep{fastvit}  &  256 & 1694 & 42.1 \\
        ResNet-101* \citep{resnetpaper}  &  224& 868  & 42.8 \\
        \rowcolor{lightergray} LowFormer-B0 (ours) &  224 & \textbf{5988}  & \textbf{44.0} \\
        \Xhline{1.0pt}
        EfficientViT-B1 \citep{efficientvit} &  224 & 2739  & 44.6 \\
        MNv4-Conv-M \citep{mobilenetv4} &  224 & 2741 & 45.3 \\
    
         \rowcolor{lightergray} LowFormer-B1 (ours) &  224 & \textbf{4237} & \textbf{45.8} \\
        
        \Xhline{1.0pt}
        
        CoaT-Lite Tiny \citep{coatlite} &  224 & 1153  & 46.4 \\
        EfficientViT-B2 \citep{efficientvit} &  224 & 1298  & 47.3 \\
        \rowcolor{lightergray} LowFormer-B1.5 (ours) &  224 & \textbf{2739} & \textbf{47.6} \\

        \Xhline{1.0pt}
        
        FastViT-SA12 \citep{fastvit} &  256 & 1075  & 48.0 \\
        CoaT-Lite Mini \citep{coatlite} &  224&  1065  & 48.3 \\
        EfficientViT-L1 \citep{efficientvit}  &  224& 1020  & 48.4 \\
        FastViT-SA36 \citep{fastvit} &  256 & 429  & \textbf{49.0} \\
        EfficientNetV2-S \citep{efficientnetv2} &  300  & 690  &\textbf{ 49.0} \\
        \rowcolor{lightergray} LowFormer-B3 (ours) &  224 & \textbf{1162}  & \textbf{49.0} \\
    
        \Xhline{1.5pt}
     
    \end{tabular}
    
    \begin{tablenotes}[para,flushleft]
        \footnotesize
        LowFormer models achieve superior speed accuracy trade-offs. Results marked with * are taken from \citep{gpr1200}. Models are grouped by mAP.
        Entries marked as bold, refer to the best results in the respective group of the table. LowFormer models are highlighted in gray
     \end{tablenotes}
  
  \end{threeparttable}

  \end{minipage} 
  \end{flushleft}
\end{table*}

\bmhead{Results}
In \Cref{tab:objectdetection}, we compare the performance of LowFormer base models in object detection against recent vision backbones.
LowFormer-B2 for example outperforms FAT-B0 \citep{fat} by \textbf{+1.0} AP, while having  93\% higher backbone throughput on resolution 512$\times$512 and only 67\% of its latency. On the other side, LowFormer-B0 with a smaller model capacity is able to achieve an increase in AP of \textbf{+1.4} compared to PVTv2-B0 \citep{pvtv2}, while 
being 4× faster in terms of TX2 latency.

In summary, LowFormer base models are able to outperform all compared vision backbones in terms of speed accuracy trade-off, when plugged into the RetinaNet framework \citep{retinanet}.

\subsubsection{Semantic Segmentation}
\label{subsec:semanseg}
We further demonstrate LowFormer's applicability to semantic segmentation in a  similar fashion as object detection. GPU throughput and TX2 latency is again measured with an input resolution of 512$\times$512 \citep{shvit, fastvit, fat}.

\bmhead{Settings}
We plug the pretrained LowFormer base models (B1,B2,B3) into the Semantic FPN framework \citep{semanticfpn}  and use the ADE20K dataset \citep{ade20k} for training and evaluation.
We train the models for 40K iterations with a batch size of 32, following \citep{fat,repvit,fastvit,efficientmodulation}. We use AdamW optimizer \citep{adamwpaper}, cosine annealing for the learning rate \citep{cosinelrpaper} with a base learning rate of $2\times 10^{-3}$ and 1K warm-up steps with linear increase. 
As evaluation metric we use mean intersection over union (mIoU) \citep{fastvit,repvit}.

\bmhead{Results}
In \Cref{tab:segmentation}, we compare the performance of LowFormer models in semantic segmentation against recent vision backbones.
LowFormer-B2 for example has 2.4× the throughput and a 25\% lower latency than EfficientFormerV2-S2\citep{efficientformerv2}, but achieves \textbf{+0.4} mIoU when plugged into Semantic FPN. 
FastViT-MA36 \citep{fastvit} achieves a similar mIoU as LowFormer-B3, but has approximately twice the TX2 latency and 35\% of its GPU throughput.

In summary, LowFormer models show significant efficiency gains compared to previous approaches, while maintaining a similar or superior mIoU.

\begin{table*}[t]
    \centering
        \caption{Evaluation of LowFormer-Track with its baseline tracker SMAT \citep{smattracker} }
    \label{tab:smattracking}

       \begin{threeparttable}[b]
       
    \resizebox{1.0\textwidth}{!}{\begin{tabular}{c|cc|cc|cc|cc|cc|cc|cc}
        \Xhline{1.5pt}
         \multirow{2}{*}{Model}  &  TX2 & GPU & \multicolumn{2}{c|}{GOT10K-val}  & \multicolumn{2}{c|}{LaSOT-Test  }  & \multicolumn{2}{c|}{TREK-150  } & \multicolumn{2}{c|}{NfS30}  & \multicolumn{2}{c|}{AVisT }  & \multicolumn{2}{c}{UAV123 }  \\
         & fps & fps & AUC & P & AUC & P & AUC & P & AUC & P & AUC & P & AUC & P \\
         \Xhline{1.0pt}

            SMAT \citep{smattracker} & \textbf{53} & 90 & 77.0 & 66.3 & 60.4 & 62.8 & 39.6 & \textbf{23.2} & 62.4 & 74.0 & 46.0 & 41.5 & 64.1 & 83.7  \\

       LowFormer-Track (ours) & 51 & \textbf{92} & \textbf{78.9} & \textbf{69.3} & \textbf{61.7} & \textbf{64.6} &\textbf{ 39.7} & 22.4 & \textbf{63.1} & \textbf{75.0} & \textbf{47.0} & \textbf{42.0} & \textbf{65.2} & \textbf{85.2} \\

     \Xhline{1.5pt}
    \end{tabular}}

      \begin{tablenotes}[para,flushleft]
        \footnotesize
        LowFormer-Track is an adaptation of the SMAT architecture that replaces the backbone with a LowFormer-B1.5 and     changes  the attention layers in the  head \\ to \lowformeratt\ layers. LowFormer-Track is similarly efficient as SMAT, but   
  consistently   outperforms  SMAT in terms of AUC and Precision, showing  the \\ benefits of the hardware efficient    design of LowFormer and  \lowformeratt. Bold values mark the best in each column
     
     \end{tablenotes}
  
  \end{threeparttable}

\end{table*}

\subsubsection{Image Retrieval}
In order to evaluate the quality of the image embedding of LowFormer, we compare it on the GPR1200 (General-Purpose Image Retrieval) benchmark \citep{gpr1200}. 

\bmhead{Settings}
The GPR1200 benchmark data is selected from several datasets, namely Google Landmarks V2, ImageNet Sketch, INat, INSTRE, SOP and IMDB Faces. The combination ensures that the data spans a variety of domains. In total GPR1200 features 12k images and  1200 different classes.
 For evaluation we follow the protocol of \citep{gpr1200} and measure performance by mean-Average-Precision (mAP). All models are executed on the input image resolution they are executed on for ImageNet evaluation. As image embedding we take the output after the final pooling, that reduces the resolution to 1×1.
We compare models by GPU throughput, as this efficiency measure directly reflects a model's ability to process large amounts of data, which is crucial for retrieving image embeddings from large datasets.
\bmhead{Results}
Results of the evaluation are depicted in \Cref{tab:gpr1200results}.
LowFormer architecture variants are able to clearly outperform all compared models in throughput accuracy trade-off (see \Cref{tab:gpr1200results}). Compared to the recently published MobileNetv4-Conv-Medium \citep{mobilenetv4}, LowFormer-B1 achieves a 0.5\% higher mAP and processes 54\% more images in the same time. Though FastViT-SA36 \citep{fastvit} for example achieves a similar mAP score as LowFormer-B3, it has less than half of the GPU throughput. 

\subsubsection{Visual Object Tracking}
Besides tasks that process images separately, we applied the LowFormer architecture to the video task of single object tracking \citep{OTB,VOT2020}. 
We use the SMAT architecture \citep{smattracker} as a baseline, replacing its backbone \citep{mobilevitv2} with LowFormer-B1.5 and substituting the attention layers in its Separable Self-Attention Head with \lowformeratt\ layers.  
We refer to this adapted model as LowFormer-Track.

\bmhead{Settings}
We train  LowFormer-Track and the baseline SMAT on the train splits of LaSOT \citep{fan2019lasot}, GOT10K \citep{GOT10k}, and COCO \citep{cocopaper}. For the latter dataset, we use data augmentations to generate image pairs from the still images, following \citep{smattracker}. 
To best compare LowFormer-Track and SMAT, we assimilate their efficiency by using LowFormer-B1.5 as a backbone, as well as train and evaluate it on a slightly lower resolution than SMAT.
For search images we resize them to 224×224 instead of 256×256, for template images we resize to 112×112 instead of 128×128. Besides that, we adopt the hyperparameter setting of SMAT for both models.
We assess both models performance across six diverse and widely used single object tracking benchmarks:  the validation set of GOT10K \citep{GOT10k}, the test set of LaSOT \citep{fan2019lasot}, the TREK-150 first person vision benchmark  \citep{matteovisobjtracking}, the NfS benchmark \citep{NfS} that predominantly contains fast-moving objects, the AVisT \citep{noman2022avist} benchmark featuring diverse scenarios with reduced object visibility, and the UAV123  benchmark \citep{UAV123} consisting of sequences captured from an aerial perspective.

\bmhead{Results}
In \Cref{tab:smattracking}, we compare LowFormer-Track and SMAT on the aforementioned single object tracking benchmarks using the Area-Under-the-Curve (AUC) and Precision (P) metric, following \cite{smattracker}. Both achieve similar fps (frames per second) on the Jetson TX2 and GPU.
However, LowFormer-Track surpasses SMAT in AUC across all benchmarks.
In terms of Precision, SMAT achieves a higher score on the TREK-150 benchmark but lags behind by a large margin in all other cases.
The superior results achieved by our adaptation of the SMAT framework highlight the significance of efficient backbone design as well as the versatility and effectiveness of our proposed \lowformeratt\ for computer vision tasks beyond image understanding.

\section{Conclusion}
In this paper, we have examined the hardware efficiency of several architectural design choices for vision backbones, such as depthwise convolutions, operating resolution of layers, fusing the MBConv block, and attention mechanisms. We have particularly shown how the execution time  differ on several different devices.
 The analysis gave us guidance for the design of a new vision backbone architecture family, named LowFormer, that features \lowformeratt, a lightweight adaptation of the original MHSA. 
The LowFormer base models (LowFormer-B0-B3) surpass competing approaches in terms of speed accuracy trade-off on GPU, the Jetson TX2, and ARM CPU. Additionally, we presented three edge GPU variants of LowFormer (LowFormer-E1/E2/E3) that  further enhance efficiency on the Nvidia Jetson TX2 and GPU.
We have shown that using LowFormer as a backbone improves efficiency across several downstream computer vision tasks, including various transfer learning image classification datasets, object detection, semantic segmentation, and image retrieval. We also presented LowFormer-Track, an adaptation of a recently published tracking framework, where we apply LowFormer-B1.5 as a backbone and incorporate our proposed \lowformeratt, clearly outperforming the baseline.
Overall, the results achieved demonstrate that the LowFormer architecture sets itself as a leading method for the implementation of efficient computer vision pipelines utilizing vision backbones.

\bmhead{Acknowledgements}{This research has been funded by the European Union, NextGenerationEU – PNRR M4 C2 I1.1, RS Micheloni. Progetto PRIN 2022 EXTRA-EYE CUP G53D23002920006, PRIN 2022 PNRR TEAM CUP G53D23006680001. 
MD received funding from the European Union’s Horizon Europe research and innovation programme under the Marie Skłodowska-Curie grant agreement n. 101151834 PRINNEVOT (CUP G23C24000910006).
This preprint has not undergone peer review (when applicable) or any post-submission improvements or corrections. The version of Record of this article is published in the international journal of computer vision (IJCV), and is available at \url{https://doi.org/10.1007/s11263-026-02873-5}.}


\bmhead{Data Availability}
ImageNet \citep{imagenet} dataset is available at \url{https://www.image-net.org/}.
GOT10K \citep{GOT10k}, LaSOT \citep{fan2019lasot}, TREK-150 \citep{matteovisobjtracking}, NfS30 \citep{NfS}, AVisT \citep{noman2022avist}, GPR1200 \citep{gpr1200}, ADE20K \citep{ade20k}, COCO 2017 \citep{cocopaper}, Oxford-IIIT-Pets \citep{petsdataset}, Oxford-102 Flowers \citep{oxfordflowers} and Stanford Cars \citep{stanfordcars} are publicly available and can be found under the references. 

\appendix


\bibliography{sn-bibliography}





\end{document}